\documentclass[sigconf]{acmart}
\AtBeginDocument{%
  }

\copyrightyear{2025}
\acmYear{2025}
\setcopyright{acmlicensed}\acmConference[CHI '25]{CHI Conference on Human Factors in Computing Systems}{April 26-May 1, 2025}{Yokohama, Japan}
\acmBooktitle{CHI Conference on Human Factors in Computing Systems (CHI '25), April 26-May 1, 2025, Yokohama, Japan}
\acmDOI{10.1145/3706598.3713217}
\acmISBN{979-8-4007-1394-1/25/04}

\begin{document}

\title[Investigating the Capabilities and Limitations of ML for Identifying Bias]{Investigating the Capabilities and Limitations of Machine Learning for Identifying Bias in English Language Data with Information and Heritage Professionals}

\author{Lucy Havens}
\email{hello@lucyhavens.com}
\orcid{0000-0001-8158-6039}
\affiliation{%
  \institution{University of Edinburgh}
  \city{Edinburgh}
  \state{}
  \country{UK}
}

\author{Benjamin Bach}
\email{benjamin.bach@inria.fr}
\orcid{0000-0002-9201-7744}
\affiliation{%
  \institution{Inria}
  \city{Bordeaux}
  \state{}
  \country{FR}
}
\affiliation{%
  \institution{University of Edinburgh}
  \city{Edinburgh}
  \state{}
  \country{UK}
}

\author{Melissa Terras}
\email{m.terras@ed.ac.uk}
\orcid{0000-0001-6496-3197}
\affiliation{%
  \institution{University of Edinburgh}
  \city{Edinburgh}
  \state{}
  \country{UK}
}

\author{Beatrice Alex}
\email{b.alex@ed.ac.uk}
\orcid{0000-0002-7279-1476}
\affiliation{%
  \institution{University of Edinburgh}
  \city{Edinburgh}
  \state{}
  \country{UK}
}


\begin{abstract}
    Despite numerous efforts to mitigate their biases, ML systems continue to harm already-marginalized people.  While predominant ML approaches assume bias can be removed and fair models can be created, we show that these are not always possible, nor desirable, goals.  We reframe the problem of ML bias by creating models to identify biased language, drawing attention to a dataset’s biases rather than trying to remove them.  Then, through a workshop, we evaluated the models for a specific use case: workflows of information and heritage professionals.  Our findings demonstrate the limitations of ML for identifying bias due to its contextual nature, the way in which approaches to mitigating it can simultaneously privilege and oppress different communities, and its inevitability.  We demonstrate the need to expand ML approaches to bias and fairness, providing a mixed-methods approach to investigating the feasibility of removing bias or achieving fairness in a given ML use case.
\end{abstract}



\received{20 February 2007}
\received[revised]{12 March 2009}
\received[accepted]{5 June 2009}

\maketitle

\section{Introduction}\label{introduction}
The Gallery, Library, Archive, and Museum (GLAM) sector collects and provides access to \textit{cultural heritage}, meaning records deemed culturally or historically significant~\cite{Smith_2006,Thomassen_2002,Welsh_2016}.  These records include audio recordings, books, letters, maps, musical instruments, paintings, and social media posts, among others.  To enable their discovery and use, librarians, archivists, and curators (``information and heritage professionals'') document them in \textit{catalogs}, writing metadata descriptions that summarize the records' cultural, economic, historical, political, social, and temporal contexts~\cite{Angel_2019,Welsh_Batley_2009}.  These descriptions shape researchers and the general public's understanding of the past because they determine how cultural heritage records can be discovered and provide an initial interpretation of the records~\cite{Schwartz_Cook_2002}.  Though for many years describing cultural heritage records was viewed as an objective process, since the late 19\textsuperscript{th} century, information and heritage professionals have increasingly recognized description as an act of interpretation and knowledge production, often undertaken in Western nations within the context of imperialism and colonialism~\cite{Duff_Harris_2002,Stoler_2002}.  Consequently, descriptions of cultural heritage records in GLAM institutions' catalogs, as well as the records themselves, reflect and reinforce power relationships within and between nations.  To counteract the marginalization of certain communities of people that catalog descriptions have reflected and reinforced, information and heritage professionals have begun dedicating resources to reviewing the descriptions for biased language, a process often referred to as \textit{critical cataloging}~\cite{Berry_2020}.    

\begin{figure*}[t]
    \centering
    \includegraphics[width=0.9\linewidth]{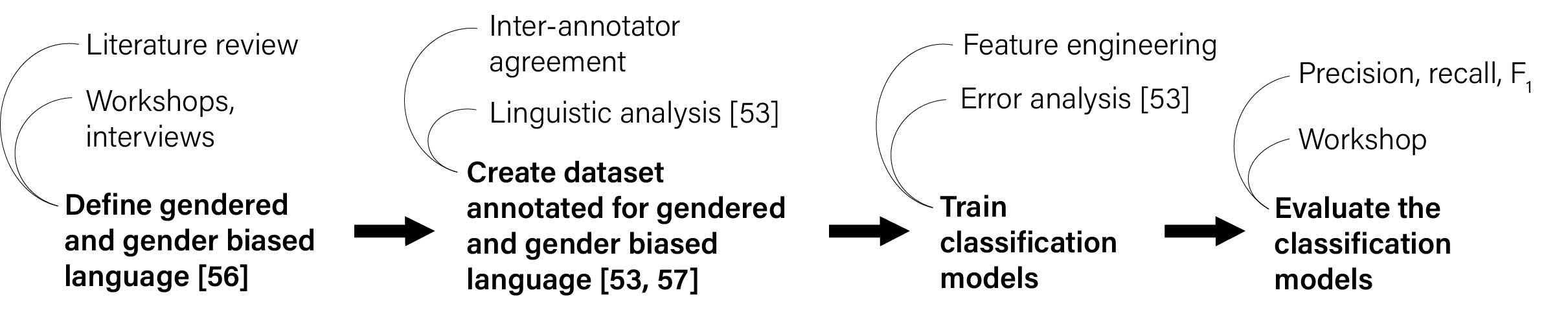}
    \caption{A flow chart of our four-step research process, with each step in boldface and the main tasks for the steps radiating out and above each step.  This paper summarizes the entire research process while focusing on the last step, ``Evaluate the classification models,'' and the substep ``Workshop,'' through which we qualitatively evaluate the models (\S\ref{workshop}).  Previous publications, as referenced in the flow chart and throughout this paper, provide more detail on the steps for defining gendered and gender biased language~\cite{Havens_2020}, creating the annotated dataset~\cite{Havens_2022}, and training and quantitatively evaluating the classification models~\cite{Havens_2024}.}
    \label{fig:research-process}
    \Description{A flow chart depicting our four-step research process.  From left to right, the steps are: (1) define gendered and gender biased language, (2) create dataset annotated for gendered and gender biased language, (3) train classification models, and (4) evaluate the classification models.  Radiating out and above the steps are two main tasks associated with each step.  For step 1, the main tasks are literature review and workshops and interviews; for step 2, inter-annotator agreement and linguistic analysis; for step 3, feature engineering and error analysis; and for step 4, precision, recall, and F1 score and workshop.}
\end{figure*}

To support critical cataloging, especially for large catalogs with millions of entries created over several centuries, Machine Learning (ML) methods increasingly seem promising.  
Recent advances in ML have led to higher performing and more readily adoptable large language models, such as OpenAI's GPT-4~\cite{OpenAI_2024}.  However, ML research and practice focus on use cases that are not relevant to information and heritage professionals in the GLAM sector for three reasons.  Firstly, predominant ML approaches to bias aim to remove bias from data or models, eliminating the possibility of analyzing bias as a historical phenomenon with ramifications for present-day society~\cite{Havens_2020,Smith_2006}.  Secondly, despite many ML publications claiming the wide generalizability of their models~\cite{Raji_2021}, GLAM publications report that state-of-the-art ML models' training datasets are too different from GLAM catalog data to make the models suitable to GLAM use cases~\cite{Lorang_2020}.  Thirdly, as we demonstrate in this paper, the types of language bias considered in ML approaches are not comprehensive enough for the GLAM sector because they typically rely on the definition and connotation of specific words, aiming to replace terms considered discriminatory with today's preferred terms, such as replacing ``transsexual'' with ``transgender.''  There are many types of bias that are more complicated to detect, however, and that have implications for both information and heritage professionals' efforts to manage bias in descriptions of cultural heritage, and for ML researchers and practitioners' approach to data and model bias (\S\ref{background}).

Given the rapidly growing  adoption of ML systems by the public, such as Google's Gemini,\footnote{\url{gemini.google.com}} and the growing integration of ML models into products, such as Microsoft's Copilot,\footnote{\url{copilot.microsoft.com}} underestimation of the biases in ML systems' underlying data threatens to magnify social injustice at an unprecedented scale.  In this paper, we combine Human-Computer Interaction (HCI) and ML methods to investigate how bias manifests in data, specifically gender bias in the language of a British GLAM catalog's descriptions.  We aim to support information and heritage professionals in managing bias in GLAM collections, and to expand ML researchers and practitioners' understandings of the varieties of bias to consider when creating and deploying ML systems.  Additionally, we aim to bridge the GLAM and ML communities, encouraging ML researchers and practitioners to draw upon information and heritage professionals' approaches and practices, because the GLAM sector has a long history of managing bias in collections at the community, national, and international scales at which ML systems are being deployed.  To that end, we asked the following research questions:
\begin{itemize}
    \item \textbf{RQ1:} How do information and heritage professionals conceptualize bias?  To what extent does this align with how ML researchers and practitioners conceptualize bias?
    \item \textbf{RQ2}: How can the identification of gender biased language with ML models inform and impact information and heritage professionals' understanding of gender biased language and management of bias in their institutions' catalogs and collections?
\end{itemize}

First, we created ML models to identify gender bias in a GLAM catalog's metadata descriptions~(\S\ref{machine-learning}).  Then, using a participatory action research approach~\cite{Swantz_2008} grounded in critical discourse analysis~\cite{Fairclough_2003} and intersectional feminism~\cite{HillCollins_2000}, we ran a workshop with information and heritage professionals to evaluate the ML models (\S\ref{workshop}). We conducted content analysis on the workshop participants' discussion, finding that (a) information and heritage professionals do conceptualize bias differently than ML researchers and practitioners, viewing bias as contextual, dynamic, and inevitable; and (b) ML models can expand information and heritage professionals' understanding of biased language and bias in their institutions' collections, though (c) ML models are limited in their ability to identify data bias comprehensively and reliably due to the contextual, dynamic, and inevitable nature of bias.  In addition to supporting information and heritage professionals' management of bias, our findings have implications for HCI researchers and practitioners working with ML systems.  Observations from the workshop~(\S\ref{observations}) highlight manifestations of bias that require information and heritage professionals' domain expertise to identify and manage, reinforcing the necessity of human-centered research methods in ML approaches to bias and fairness.  Our findings also have implications for ML researchers and practitioners, surfacing complexities of data bias that have been overlooked and demonstrating that, in certain use cases, removing bias and creating fair ML systems may not be possible nor desirable.  Our mixed-methods evaluation of ML models and their underlying data provides an approach to investigating the feasibility of removing bias or achieving fairness in a given ML use case.

\section{Background}\label{background}

Existing work in the GLAM sector acknowledges how societal power relationships shape what GLAM institutions collect, how those collections are described, and how these processes shape cultural and historical narratives that produce and perpetuate social biases~\cite{Cook_2011,Duff_Harris_2002}.  For example, information and heritage professionals have critiqued how these processes have enacted privilege and oppression based on gender~\cite{Olson_2001}, sexuality~\cite{Adler_2017}, race~\cite{Furner_2007}, and colonialism~\cite{Stoler_2002}.  Existing GLAM work also acknowledges the power that information and heritage professionals have to reject and reshape narratives that portray inaccurate and harmful views of particular communities~\cite{Caswell_2022,Caswell_Cifor_2016}.  Aiming to empower historically marginalized communities, information and heritage professionals have been reviewing and revising the catalog metadata descriptions that summarize and contextualize their GLAM institutions' collections.  Such anti-oppressive, critical cataloging work has been largely manual~\cite{Berry_2020,Tai_2021}, requiring GLAM institutions to divide their already limited resources between describing new collections and revisiting existing descriptions.

ML affords efficiency and scalability that appeals to information and heritage professionals because it would advance the profession's primary goal of providing access to information~\cite{Angel_2019}.  The ability to more quickly describe heritage could help address the backlog of items that have not been or were only minimally cataloged~\cite{Greene_Meissner_2005,Trace_2022}.  The Library of Congress, for instance, adds approximately 10,000 new items to its collections every day and employs approximately 3,200 people, only a fraction of whom are catalogers responsible for describing incoming items~\cite{LC,LC2}.  Though experiments with automated metadata generation are underway~\cite{LC_Labs,Jaillant_2022}, GLAM institutions still rely on information and heritage professionals' manual review and description of items to ensure their discoverability for the public.  Existing research identifies three factors limiting ML methods and models' applicability to GLAM use cases: (1) a domain mismatch between state-of-the-art ML models' training data and GLAM catalog data~\cite{Cordell_2020,Havens_2025b,Lorang_2020}, (2) a skills gap between typical information and heritage professionals' education and training and the data and computer science education and training needed to create and use ML models~\cite{Havens_2024,Marciano_2022,Terras_2018}, and, most importantly for this paper, (3) a mismatch between conceptualizations of bias and fairness between information and heritage professionals and ML researchers and practitioners~\cite{Coleman_2020,Posner_2016}.  Research demonstrating the inefficacy of ML approaches to minimizing bias~(\S\ref{related-work-ml}) justify information and heritage professionals' cautious approach to adopting pre-trained ML models.

Due to the complexity of social biases, which results from multifaceted and ever-changing societal power relationships ~\cite{Crenshaw_1989,Crenshaw_1991,HillCollins_2000,Young_2011}, we focus our work on gender bias, working with British English text from a Scottish Archive's catalog metadata descriptions.  We adopt our previously published definition of gender biased language as: \textit{``language that creates or reinforces inequitable power relations among people, harming certain people through simplified, dehumanizing, or judgmental words or phrases that restrict their gender identity; and privileging other people through words or phrases that favor their gender identity''}~\cite[p.~108]{Havens_2022}.  We are interested in how information and heritage professionals can provide insight on implicit ways that gender bias manifests in language, as opposed to the explicit forms of biased language more widely studied in ML: primarily terminology that is derogatory or stereotypical for content moderation and machine translation use cases~\cite{Ciora_2021,Schmidt_Wiegand_2017}.  ML publications on gender bias have reported omissions of women in training data~\cite{Buolamwini_Gebru_2018}, stereotypical representations of men and women in models~\cite{Noble_2018}, and omissions and misrepresentations of people of trans and gender diverse identities in models~\cite{Keyes_2018}.  Research on sexism in computing has demonstrated the socio-technical nature of bias, with gender biased social structures resulting in datasets that encode gender biases and models that have gender biases engineered into them~\cite{Hicks_2018,Hicks_2021,Perez_2019}.  These patterns in ML reflect patterns in GLAM: cultural heritage records about women and people of trans and gender diverse identities have been under-collected~\cite{Hessel_2023,Shopland_2020} and, when collected, under-documented and misrepresented~\cite{Olson_2001}, perpetuating gender stereotypes and under-acknowledging the contributions of these gender communities~\cite{Beard_2017}.  Such omission and misrepresentation has wide-reaching consequences for women and people of trans and gender diverse identities, affecting their experiences in everything from transportation to medical care to work opportunities~\cite{Bohannon_2023,Dunsire_2018,Perez_2019}.

The challenge with addressing gender biased language is that it is subjective and contextual.  The same words or phrases may be interpreted differently by different people and may be harmful in one context but not another, such as varying attitudes towards the term ``queer'' and changes in the term's connotation over time in Western cultures~\cite{Bucholtz_1999,Schulz_1975,Shopland_2020}.  Consider the gender bias in this description of Dr. Susan Binnie Anderson (1898-1975): 
\begin{quote}
    \textit{As a young married woman with two small children--Kathleen and Margaret--she found it difficult to keep her medical career going [...] She twice tried to return to medicine full-time [sic] but this just wasn't to be.\footnote{\url{archives.collections.ed.ac.uk/repositories/2/resources/546}}}  
\end{quote}
There is no derogatory terminology or explicitly stereotypical word choice in the description.  Yet, one could argue the description communicates an implicit assumption that the state of being young, married, a woman, and a mother negates the possibility of a medical career.  Patriarchal societal structures during Dr. Anderson's lifetime may have caused the difficulty she faced, but the description does not consider this.  On the other hand, the description does not provide information about Dr. Anderson's husband, who may have also found it difficult to keep his career going being young, married, and a father, in which case the difficulty of maintaining a career would be implicitly attributed to the challenges of parenthood.  Some people may identify gender bias in this description while others may not.  This implicit type of bias and the uncertainty surrounding it is understudied in ML research and, with the growing adoption of generative ML models for search, summarization, writing, and decision making, of increasing concern~\cite{Hofmann_2024,Lai_2023,Lucy_2021}.

\section{Related Work}\label{related-work}
    
    \subsection{Bias in ML}\label{related-work-ml}
    ML approaches to data and model bias predominantly view bias as a technical flaw that needs to be mitigated in order to create fair ML systems (for summaries, see \cite{CorbettDavies_2023,Hort_2024,Mehrabi_2021,Pessach_2022,Sun_2019}).  Approaches to creating unbiased, fair ML systems include efforts to measure bias in datasets~\cite{Basta_2020,Kabir_2024} and models~\cite{Jentzsch_2022,Jin_2021}, create balanced datasets~\cite{Webster_2018,Zhao_2018_Debias} and create or fine-tune models to treat different demographic groups fairly~\cite{Krasanakis_2018,Samorani_2022}.  Several scholars have noted that predominant approaches to ML bias and fairness rely on quantitative analyses and mathematical representations, failing to account for the societal structures that cause and perpetuate social biases~\cite{Baumer_2017,Hoffmann_2019,Kasirzadeh_2022,McCradden_2020}.  The work of Rodolfa \textit{et al.}~\cite{Rodolfa_2020}, Ciora \textit{et al.}~\cite{Ciora_2021}, and Samorani \textit{et al.}~\cite{Samorani_2022} illustrate how considering the distribution of power in the societal context in which an ML system will be deployed can inform the system's design.  However, the focus of these scholars' work was creating ML systems that do not behave in a biased manner.  More research is needed on use cases where biases are inevitable or full automation is not desirable, such as in healthcare, criminal justice, and the GLAM sector.

    Despite the proliferation of ML bias mitigation methods put forth, reliable, generalizable mitigation methods have yet to be found.  Research inspecting the efficacy of bias mitigation approaches has yielded inconsistent results, with some findings indicating successful mitigation~\cite{Jin_2021} and others indicating no conclusive impact~\cite{Goldfarb-Tarrant_2021_Metrics,Orgad_Belinkov_2022,Lauscher_2019}.  Most concerning are findings that indicate existing approaches have amplified bias~\cite{Hofmann_2024} and become new causes of harm to the marginalized communities that the approaches were intended to help~\cite{CorbettDavies_2023,DiasOliva_2021}.  While recently proposed methods of matched guise probing~\cite{Hofmann_2024} and mechanistic interpretability~\cite{Kastner_Crook_2024,Liberum_2024} offer promising approaches for more effectively detecting bias, understanding how biases are engineered into models and comprehensively testing for them are ongoing challenges.  This may be due to manifestations of bias that are not well-understood, such as the covert forms of sexism Ciora \textit{et al.} investigate for machine translation in Turkish~\cite{Ciora_2021} and covert forms of racism that Hofmann \textit{et al.} found in large language models~\cite{Hofmann_2024}.  This may also be due to overconfidence in the generalizability of datasets, models, benchmarks, and metrics~\cite{Baumer_2017,Paullada_2021,Raji_2021,Wagstaff_2012}.  Our work aims not only to better understand the limitations of ML approaches to bias, but also to demonstrate how to test for those limitations.

    To improve ML approaches to bias and fairness, scholars have called for greater interdisciplinary research and context-specific approaches~\cite{Blodgett_2020,CorbettDavies_2023}.  This has led to ML research that draws upon sociolinguistics~\cite{Cao_2020}, feminism~\cite{Klein_D’Ignazio_2024}, legal and political concepts~\cite{Barocas_Selbst_2016,Kasirzadeh_2022}, and human-centered research methods common to the HCI community~\cite{Nekoto_2020,Wang_2021,Lai_2023,Markl_Lai_2021}, among others.  HCI scholars have grown the fields of human-centered ML and eXplainable Artificial Intelligence (XAI), using surveys, interviews, and participatory research methods to study ML systems in specific contexts~\cite{HCRAI_2023,Liao_2022,Shneiderman_2022}.  This work has distinguished the needs of different end users in different usage contexts~\cite{Dhanorkar_2021,Liao_2020} and shed insight on users' perceptions of ML systems, including perceptions of bias and fairness, and the implications of these perceptions for users' workflows~\cite{Holstein_2019,Nourani_2024,Sambasivan_2021,Yuan_2023}. Still, research is needed to investigate how an ML system's end users conceptualize bias beyond the confines of an ML system; that is the focus of this paper.
    
    Although the GLAM sector has a longer history of managing bias than the ML community, few ML researchers or practitioners have drawn upon GLAM approaches for guidance on working with data.  Jo and Gebru encourage ML researchers and practitioners to look to archival and library cataloging practices for guidance on dataset documentation~\cite{Jo_2020}.  Thylstrup \textit{et al.} use archives as a metaphor for big data, exploring the questions, challenges, and opportunities that arise when approaching large-scale datasets as archival collections~\cite{Thylstrup_2021}.  Crawford and Paglen note that information and heritage professionals have long recognized the political implications of classification in a comparison of ImageNet's classification scheme with a historical version of the Library of Congress Classification scheme~\cite{Crawford_Paglen_2019}.  Yet there are also GLAM approaches with more practical applicability to ML.

    \subsection{Bias in GLAM}
    
    GLAM approaches to bias offer insight on why efforts to mitigate bias in ML systems continue to fall short.  Given the predominance of manual data work in the GLAM sector, information and heritage professionals are particularly attune to the implications of inequitable societal power relationships on creating, collecting, organizing, analyzing, and using data~\cite{Bowker_Star_2000,Caswell_2022,Duff_Harris_2002,Hessel_2023,Ortolja-Baird_Nyhan_2022,Stoler_2002}.  They have looked to critical theories to develop new practices that minimize harms from biases in GLAM catalogs and the collections they describe.  Nonetheless, information and heritage professionals' conceptualizations of bias have not been characterized in a way that is easily translatable to training and evaluating ML models.  In this paper, by evaluating ML models with information and heritage professionals (\S\ref{workshop}), we are able characterize bias in a manner that draws upon GLAM domain expertise and is translatable to ML system creation (\S\ref{discussion}).

\begin{table*}[t]
    \caption{The coding taxonomy of our training dataset's source data, reproduced with authors' permission~\cite{Havens_2022}.  In the far right column, coded text is boldfaced and, for the four middle rows, co-referents determining the coding of a name are italicized.}\label{t:taxonomy}
    \begin{tabular}{p{0.14\linewidth}p{0.11\linewidth}p{0.22\linewidth}p{0.43\linewidth}}
    \toprule
    Code &
      Code Category &
      Explanation &
      Example Description (Coded Text in Boldface) \\ \midrule
    Generalization &
      Linguistic &
      Gender-specific term used to refer to a group that could include more than one gender &
      His classes included Anatomy, Practical Anatomy, \textbf{Midwifery} and Diseases of Women. \\
    Gendered Role &
      Linguistic &
      Grammatically- or lexically-gendered word &
      New map of Scotland for \textbf{Ladies} Needlework. \\
    Gendered Pronoun &
      Linguistic &
      Feminine, masculine, or non-binary (neo-) pronouns &
      \textbf{He} obtained surgical qualifications in 1873. \\ \midrule
    Feminine &
      Person Name &
      Person referred to with a grammatically or lexically feminine term &
      Although \textbf{Yolanda Sonnabend} shared the copyright of this book with Waddington, \textit{she} was not credited as an illustrator. \\
    Masculine &
      Person Name &
      Person referred to with a grammatically or lexically masculine term &
      \textbf{Martin Luther}, the \textit{man} and \textit{his} work. \\
    Non-binary &
      Person Name &
      Person referred to with a grammatically or lexically non-binary term &
      \textbf{Francis McDonald} went to the University of Edinburgh where \textit{they} studied law. \\
    Unknown &
      Person Name &
      Person referred to without a grammatically or lexically gendered term &
      Testimonials in favor of \textbf{Niecks}. \\ \midrule
    Occupation &
      Contextual &
      Job title &
      He became a \textbf{surgeon} with the Indian Medical Service. \\
    Omission &
      Contextual &
      Exclusion of a person’s identity or contribution &
      This group portrait of Laurencin, Apollinaire, and Picasso and \textbf{his mistress} became the theme of a larger version. \\
    Stereotype &
      Contextual &
      Reductive or derogatory expectation of a person’s behaviors or preferences &
      \textbf{Jewel took an active interest in her husband's work}, accompanying him when he travelled, sitting on charitable committees, looking after missionary furlough houses and much more. \\ \bottomrule
    \end{tabular}
\end{table*}
    
    Critical theories originate in social justice movements and interdisciplinary research, emphasizing context-specific approaches that consider power and social and cultural reproduction~\cite{Kushner_Morrow_2003,Leavy_2017}.  Critical heritage studies scholar Smith draws on \textit{critical discourse analysis}~\cite{Smith_2006}, which views the meaning of language as a result of the words used and the context in which language is produced and received~\cite{Fairclough_2003,Bucholtz_2003}.  The former is referred to as the ``internal relations'' of a text; the latter, the ``external relations''~\cite{Fairclough_2003}.  Smith draws on critical discourse analysis to expand conceptualizations of heritage beyond the predominant Western idea of large, tangible artifacts, defining heritage as a process of creating, sharing, using, and recreating knowledge~\cite{Smith_2006}.  We extend her framework to data, considering how internal and external relations of an ML model's training data affect the biases of a model.  By starting with biased language in the context of ML systems, existing ML research focuses on the internal relations of data, such as explicitly derogatory language~\cite{Sahoo_2022} or pornographic images~\cite{Birhane_Prabhu_2021}.\footnote{There is ML research on bias that considers external relations such as how skews in research participants' demographic characteristics can bias research findings~\cite{Linxen_2021} and how gender biases in language relate to the person producing, receiving, or being described in text~\cite{Dinan_2020}.  However, the manifestation of bias being studied in this area of work is about agreements and disagreements between research participants or data labels.  Moreover, in this area, researchers often view disagreeing labels an an indication of different interpretations of text~\cite{Pang_2023}.  We position our work among research that studies how bias manifests in language and allows for the possibility that disagreeing labels may indicate different opinions about text that has, in fact, been interpreted to mean the same thing~\cite{Basile_2021,Goree_Crandall_2023}.  See \cite{Havens_2025b} for a broader consideration of bias in GLAM catalogs and collections.}  We expand upon this work by starting from human understandings of social biases, or the external relations of a dataset, facilitating a discussion about bias with our training data's domain experts: information and heritage professionals (\S\ref{workshop}).  This enables us investigate which types of biases ML models cannot reliably identify or avoid reinforcing.
    
    Considering data's external relations aligns with \textit{feminism}, a critical theory that rejects the idea of a universal truth, instead viewing knowledge as situated~\cite{Haraway_1988,Harding_1995}.  Information and heritage professionals have drawn upon feminism for managing bias in GLAM collections~\cite{Caswell_2022, Caswell_Cifor_2016,Olson_2001}.  Havens draws on intersectional feminism~\cite{HillCollins_2000,Crenshaw_1989,Crenshaw_1991} to inform GLAM description practices, including critical cataloging~\cite{Havens_2025a}.  She proposes using Hill Collins' matrix of domination~\cite{HillCollins_2000} when analyzing the external relations that influence the meaning of catalog metadata descriptions and how biases manifest in their language.  The matrix of domination consists of four interacting domains through which power is exercised in society: the \textit{structural} domain organizes oppression and privilege (\textit{e.g.}, a government body), the \textit{disciplinary} domain manages oppression and privilege (\textit{e.g.}, a court), the \textit{hegemonic} domain justifies oppression and privilege (\textit{e.g.}, a museum), and the \textit{interpersonal} domain is where oppression and privilege are experienced (\textit{e.g.}, by a person).  The matrix of domination provides a structure around which to conceptualize power relationships between individuals, communities, and organizations that create social biases, oppressing some and privileging others.  We used critical discourse analysis and intersectional feminism as lenses to guide our workshop analysis, through which we sought to understand why ML methods alone are insufficient for understanding how bias manifests in data (\S\ref{observations}).

\section{Machine Learning}\label{machine-learning}
We use ML to reframe the problem of bias.  Our use of ML aligns with concepts of ``sandcastling''~\cite{Hinrichs_2019} and ``doing design otherwise''~\cite{DiSalvo_2022}, which emphasize the process of building something.  Hinrichs \textit{et al.} use the term ``sandcastle'' to describe data visualizations built for digital humanities projects that ``facilitate a collaborative discourse about the means through which information is developed, collected, and represented''~\cite[p.~i86]{Hinrichs_2019}.  DiSalvo describes ``doing design otherwise'' as the use of design methods to interrogate problems rather than create solutions to problems, engaging in social and political efforts in the process~\cite{DiSalvo_2022}.  For both sandcastling and doing design otherwise, what is built is situated, fragile, and incomplete; the research value is in the questions and insights gleaned through the critical thinking and collaborative, cross-disciplinary analysis that the building process surfaces.  The fragile and incomplete nature of what is built encourages cross-disciplinary participation, discussion, and critique~\cite{DiSalvo_2022,Hinrichs_2019}.  As such, the built thing has a purpose similar to that of a technology probe~\cite{Hutchinson_2003} or boundary object~\cite{Star_1989}.  The ML models we develop and use in this paper function more as sandcastles than as algorithmic innovations, enabling us to facilitate a discussion with GLAM domain experts to surface questions and insights about manifestations of gender bias that ML models can and cannot identify (\S\ref{workshop}).  Moreover, given growing evidence of how ML innovations are reinforcing the power of the world's most privileged people rather than having a democratizing impact on global society~\cite{Birhane_2020,Crawford_2021,GrayWidder_2023,Hearn_2024,Hicks_2018}, our use of ML upends this trend by heightening the visibility of data bias.  We use ML to call attention to inequitable gender-based power relationships, informing and motivating efforts to empower women and people of trans and gender diverse identities.
    
    \subsection{Training Dataset}\label{training-dataset}    
    We created our ML models' training dataset from our previously published, human-coded datasets of descriptions from the archival catalog of the University of Edinburgh's Heritage Collections~\cite{Havens_2022}.  The catalog is organized according to the General International Standard for Archival Description (ISAD(G))~\cite{isadg}.  The dataset's descriptions came from four metadata fields: ``Title,'' ``Scope and Contents,'' ``Biographical / Historical,'' and ``Processing Information.''  The descriptions document heritage in a range of formats (\textit{e.g.}, letters, photographs, degree certificates) covering a variety of topics (\textit{e.g.}, religion, research, teaching, architecture).  The coding taxonomy has ten types of gendered and gender biased language: \textit{\textbf{Gendered Pronoun}}, \textit{\textbf{Gendered Role}}, \textit{\textbf{Generalization}}, \textit{\textbf{Feminine}}, \textit{\textbf{Masculine}}, \textit{\textbf{Non-binary}}, \textit{\textbf{Unknown}}, \textit{\textbf{Occupation}}, \textit{\textbf{Omission}}, and \textit{\textbf{Stereotype}} (Table \ref{t:taxonomy}).  The dataset contains 11,888 descriptions, totaling 399,957 words across 24,474 sentences, with a total of 55,260 codes.  Only 9 of the codes were applied; \textit{\textbf{Non-binary}} is not present in the dataset.  \ref{a:ml-preprocess} details our preprocessing work to transform the dataset for model training and testing.  For linguistic analysis of the coded data, see \cite[p.~86--93]{Havens_2024}.

    \subsection{Model Design, Training, and Evaluation}\label{text-classification}
    
    Given our interest in supporting information and heritage professionals' understanding of gender bias in a specific GLAM catalog, we created text classification models in a supervised, traditional ML setup.  This ensured that the gender biases the models identified originated in our training dataset.  Though deep learning models are considered state-of-the-art, they pose a risk to the validity of our research: biases from a pre-trained, or foundation, model's initial training dataset would be injected into our customized version of the model~\cite{Goldfarb-Tarrant_2021_Metrics,Ladhak_2023,Steed_2022}.  ML researchers and practitioners have yet to develop an approach to distinguishing biases in a pre-trained model's training dataset from biases in the dataset used to customize that model.  Since we aimed to identify gender biased language in our training dataset only, we did not use pre-trained deep learning models, because their classification of the descriptions would have been influenced by biased language in other data.  Given the unique task of our models, namely, classifying text as \textbf{\textit{Omission}} or \textbf{\textit{Stereotype}} based on the definitions in Table \ref{t:taxonomy}, the models we report in this paper provide a baseline against which future ML experiments can be compared.  In the remainder of this section, we briefly summarize the training and testing process for our classification models; \S\ref{a:ml-design} further details the process for reproducibility.
    
    We created three types of ML models.  The Linguistic Classifier (LC) is a multilabel token classification model that codes grammatically or lexically gendered words, classifying text as \textit{\textbf{Gendered Pronoun}}, \textit{\textbf{Gendered Role}}, and \textit{\textbf{Generalization}}.  The Person Name and Occupation Classifier (PNOC) is a multiclass sequence classification model coding phrases, sentences, or paragraphs based on the grammatical or lexical gender of a person's co-referents (see the right-most column of Table \ref{t:taxonomy} for examples), classifying text as \textit{\textbf{Feminine}}, \textit{\textbf{Masculine}}, and \textit{\textbf{Unknown}};\footnote{The \textbf{\textit{Non-binary}} code does not appear in the training data so it could not be used in the models.} and coding job titles, classifying text as \textit{\textbf{Occupation}}.  The Omission and Stereotype Classifier (OSC) is a multilabel document classification model that codes descriptions as gender biased, classifying text as \textit{\textbf{Omission}} and \textit{\textbf{Stereotype}}.

\begin{table}[b]
    \caption{Macro performance scores for each Omission and Stereotype Classifier (OSC), the average of the human coders' Inter-Annotator Agreement (IAA) scores, and the average of the five human coders' agreement scores with the training dataset.  Human coder scores are reproduced with permission~\cite{Havens_2022}.}\label{t:osc-scores}
    \begin{tabular}{p{0.3\linewidth}p{0.15\linewidth}p{0.15\linewidth}p{0.15\linewidth}}
    \toprule
    & Precision & Recall & F\textsubscript{1} \\ \midrule
    Baseline OSC        & 0.899   & 0.643   & 0.747 \\
    Cascade 1  OSC       & 0.896   & 0.644   & 0.747 \\
    Cascade 2  OSC       & 0.889   & 0.667   & 0.760 \\
    Cascade 3  OSC       & 0.888   & 0.669   & 0.761 \\ \midrule
    Coders' IAA      & 0.485   & 0.420   & 0.427 \\ 
    Coders vs. Train & 0.996   & 0.662   & 0.786 \\ \bottomrule
    \end{tabular}
\end{table}

    In addition to creating these models to interrogate conceptualizations of gender biased language, we were interested in investigating correlations between grammatically and lexically gendered language and gender biased language.  For instance, would identifying the presence of gendered roles in a description (\textit{e.g.}, ``Lady,'' ``father,'' ``Mrs.'') help a classifier determine whether or not a description communicates a gender stereotype?  To investigate this, we trained baseline versions (meaning no feature engineering) of the LC, PNOC, and OSC, and then we sequentially combined these classifiers in cascades, using a previous classifier's predictions as features for the subsequent classifier in a cascade, resulting in three feature-engineered versions of the OSC.  The cascades are:
    \begin{itemize}
        \item \textbf{Cascade 1: LC > PNOC > OSC:} We ran the LC and then passed its predictions (\textit{\textbf{Gendered Pronoun}}, \textit{\textbf{Gendered Role}}, and \textit{\textbf{Generalization}} codes) to a PNOC as features.  Then, we ran that PNOC and passed its predictions (\textit{\textbf{Feminine}}, \textit{\textbf{Masculine}}, \textit{\textbf{Unknown}}, and \textit{\textbf{Occupation}} codes) and the LC's predictions to an OSC as features.
        
        \item \textbf{Cascade 2: LC > OSC:} We ran the LC and then passed its predictions (\textit{\textbf{Gendered Pronoun}}, \textit{\textbf{Gendered Role}}, and \textit{\textbf{Generalization}} codes) to an OSC as features. 
        
        \item \textbf{Cascade 3: PNOC > OSC:} We ran a PNOC and then passed its predictions (\textit{\textbf{Feminine}}, \textit{\textbf{Masculine}}, \textit{\textbf{Unknown}}, and \textit{\textbf{Occupation}} codes) to an OSC as features.
    \end{itemize}
    By comparing the performance scores of the baseline and feature-engineered OSCs, we can evaluate whether including gendered language codes as features improves an OSC's ability to classify text as gender biased.

\begin{table*}[ht]
    \caption{Comparing human coders and text classification models' performance for \textbf{\textit{Omission}} and \textbf{\textit{Stereotype}}.  The ``Coders' IAA'' column reports the average of the IAA scores between the human coders~\cite{Havens_2022}.  The ``Coders vs. Train'' column displays the average of the human coders' agreement scores with the training dataset.  The ``OSC'' columns report the performance scores of Omission and Stereotype Classifiers.  Scores are calculated loosely, meaning two codes are considered to be in agreement if applied to overlapping or exactly matching text spans.  The highest scores per label are in bold.  Human coder scores are reproduced with permission~\cite{Havens_2022}.}\label{t:osc-label-scores}
    \begin{tabular}{p{0.1\linewidth}|p{0.07\linewidth}p{0.07\linewidth}p{0.07\linewidth}|p{0.07\linewidth}p{0.07\linewidth}p{0.07\linewidth}|p{0.07\linewidth}p{0.07\linewidth}p{0.07\linewidth}}
        \toprule
        & \multicolumn{3}{c|}{Coders' IAA} & \multicolumn{3}{c|}{Coders vs. Train} & \multicolumn{3}{c}{Baseline OSC} \\ \midrule
        Label & Precision & Recall & F\textsubscript{1} & Precision  & Recall   & F\textsubscript{1} & Precision  & Recall   & F\textsubscript{1} \\ \midrule
        Omission   & 0.525 & 0.400 & 0.429      & \textbf{0.996} & \textbf{0.631} & \textbf{0.757}     & 0.859 & 0.539 & 0.663 \\
        Stereotype & 0.446 & 0.441 & 0.426      & \textbf{0.995} & 0.694 & 0.815              & 0.939 & 0.746 & 0.832 \\ \bottomrule
    \end{tabular}
\end{table*}
\begin{table*}
    \begin{tabular}{p{0.1\linewidth}|p{0.07\linewidth}p{0.07\linewidth}p{0.07\linewidth}|p{0.07\linewidth}p{0.07\linewidth}p{0.07\linewidth}|p{0.07\linewidth}p{0.07\linewidth}p{0.07\linewidth}}
        \toprule
        & \multicolumn{3}{c|}{Cascade 1 OSC} & \multicolumn{3}{c|}{Cascade 2 OSC} & \multicolumn{3}{c}{Cascade 3 OSC} \\ \midrule
        Label & Precision & Recall & F\textsubscript{1} & Precision & Recall & F\textsubscript{1}  & Precision & Recall & F\textsubscript{1} \\ \midrule
        Omission   & 0.853 & 0.546 & 0.666      & 0.848 & 0.563 & 0.676            & 0.845 & 0.571 & 0.682 \\
        Stereotype & 0.940 & 0.741 & 0.829      & 0.930 & \textbf{0.771} & \textbf{0.843}   & 0.932 & 0.766 & 0.841 \\ \bottomrule
    \end{tabular}
\end{table*}
    
    All models were trained and tested using a modified five-fold cross-validation approach (see \S\ref{a:ml-design} for detail).  We evaluated the models using precision, recall, and F\textsubscript{1} scores, optimizing for F\textsubscript{1} score (\S\ref{ml-models}).

    \subsection{Model Performance}\label{ml-models}
    
    In this section we summarize the results of our ML experiments most relevant to the workshop (\S\ref{workshop}): the classification of gender biased language in descriptions with the \textbf{\textit{Omission}} and \textbf{\textit{Stereotype}} codes.  \S\ref{a:ml-results} reports on model performance for the remaining codes in the taxonomy.  For further error analysis and discussion of the classification results, see \cite[p.~149--166]{Havens_2024}.
    
    The OSCs' F\textsubscript{1} scores suggest that coding tokens as grammatically or lexically gendered improves a model's ability to classify gender biased language because the feature-engineered OSCs in cascades 1-3 outperform the baseline OSC (Table \ref{t:osc-scores}).  The OSCs' performance scores exceed the human coders' IAA overall (across both codes) and per \textbf{\textit{Omission}} and \textbf{\textit{Stereotype}} code.  The OSCs' recall and F\textsubscript{1} scores also exceed the human coders' agreement with the training dataset for the \textbf{\textit{Stereotype}} code, but not overall, due to the OSCs' lower performance classifying text with the \textbf{\textit{Omission}} code. Given that the training dataset has over twice as many \textbf{\textit{Omission}} codes (4,032) than \textbf{\textit{Stereotype}} codes (1,601), the greater difficulty classifying descriptions with \textbf{\textit{Omission}} than with \textbf{\textit{Stereotype}} is surprising.  However, the OSCs' lower scores for \textbf{\textit{Omission}} aligns with human coders' scores, which were also lower for \textbf{\textit{Omission}}.  These results suggest that there may be greater variation in the language coded as \textbf{\textit{Omission}} than as \textbf{\textit{Stereotype}}.
    
    Though the OSCs' performance scores are low relative to state-of-the-art classification models that rely on deep learning methods, the low human coder IAA scores indicate that classifying descriptions as \textbf{\textit{Omission}} and \textbf{\textit{Stereotype}} is a very difficult task.  Our classifiers are the first ML models built to identify manifestations of gender biased language in a GLAM catalog's metadata descriptions, so they provide a baseline upon which future work can build (\S\ref{limitations}).  The models' performance demonstrates the viability of training ML models on GLAM catalog metadata, which has been questioned in previous work~\cite{Lorang_2020}.
        
    Our model design pushes back against predominant approaches to creating and evaluating ML systems.  Model performance metrics typically do not consider bias or fairness, and the identification of bias in a model does not affect its status as ``state-of-the-art;'' consider the continued widespread use of GPT-4 despite Hofmann \textit{et al.}'s findings of racial bias in the model~\cite{Hofmann_2024}.  Given the misrepresentations and omissions that result from biases in ML systems, we argue that performance evaluations should, as a standard, include measures of bias and fairness.  By creating ML models trained to classify gender biased language, we unavoidably tie bias measurement to performance evaluation.
    
    The next section reports our human-centered ML evaluation: a discussion of our models' training data and output classifications in a workshop.  Our workshop participants (\S\ref{workshop}) are information and heritage professionals who work in the University of Edinburgh's Heritage Collections, so their knowledge of the collections represented in our training dataset enable them to provide an informed evaluation of the ML models' classifications, comparing and contrasting their interpretation of a description's gender bias with the models' interpretation.  By creating ML models to classify descriptions rather than manually coding descriptions for gender biased language ourselves, we avoid coding descriptions in a manner that emphasizes only the biases of which we as authors are most cognizant (the classifiers were trained on data that aggregates five expert annotators' coding~\cite{Havens_2022}).  Additionally, creating the models enabled us to quickly classify more descriptions than we could have coded manually, providing the data needed for workshop
    activity 2.  Our models-as-sandcastles thus have two purposes: (a) for information and heritage professionals, illustrating how ML could enable more efficient, large-scale reviews of GLAM catalog descriptions for bias; (b) for us as researchers, motivating cross-disciplinary discussion that provides insights on the capabilities and limitations of ML for identifying biased language in a real-world use case.  As we describe in \S\ref{discussion}, these insights have implications for all ML researchers and practitioners, providing guidance on how to improve dataset curation, data coding, and model design processes.

\section{Workshop}\label{workshop}

    \subsection{Setup and Procedure}\label{workshop-setup}
   
    We facilitated an open-format workshop to evaluate the ML models with 10 information and heritage professionals.  We wrote pre-defined questions to guide participants' discussion during the workshop in the manner of a semi-structured interview~\cite{Ritchie_Lewis_2003}, which we refined during a pilot workshop with three doctoral students and one postdoctoral researcher in digital humanities and information science.  The workshop was approved by the ethics board of the School of Informatics at the University of Edinburgh (reference 2019/81479).  The workshop discussion was audio-recorded with participants' consent using a single microphone connected to Zoom.  We corrected the Zoom-generated transcript of the workshop manually and shared the transcript with participants to ensure its accuracy prior to analyzing the discussion.

\begin{figure*}[ht!]
    \centering
        \includegraphics[width=0.75\linewidth]{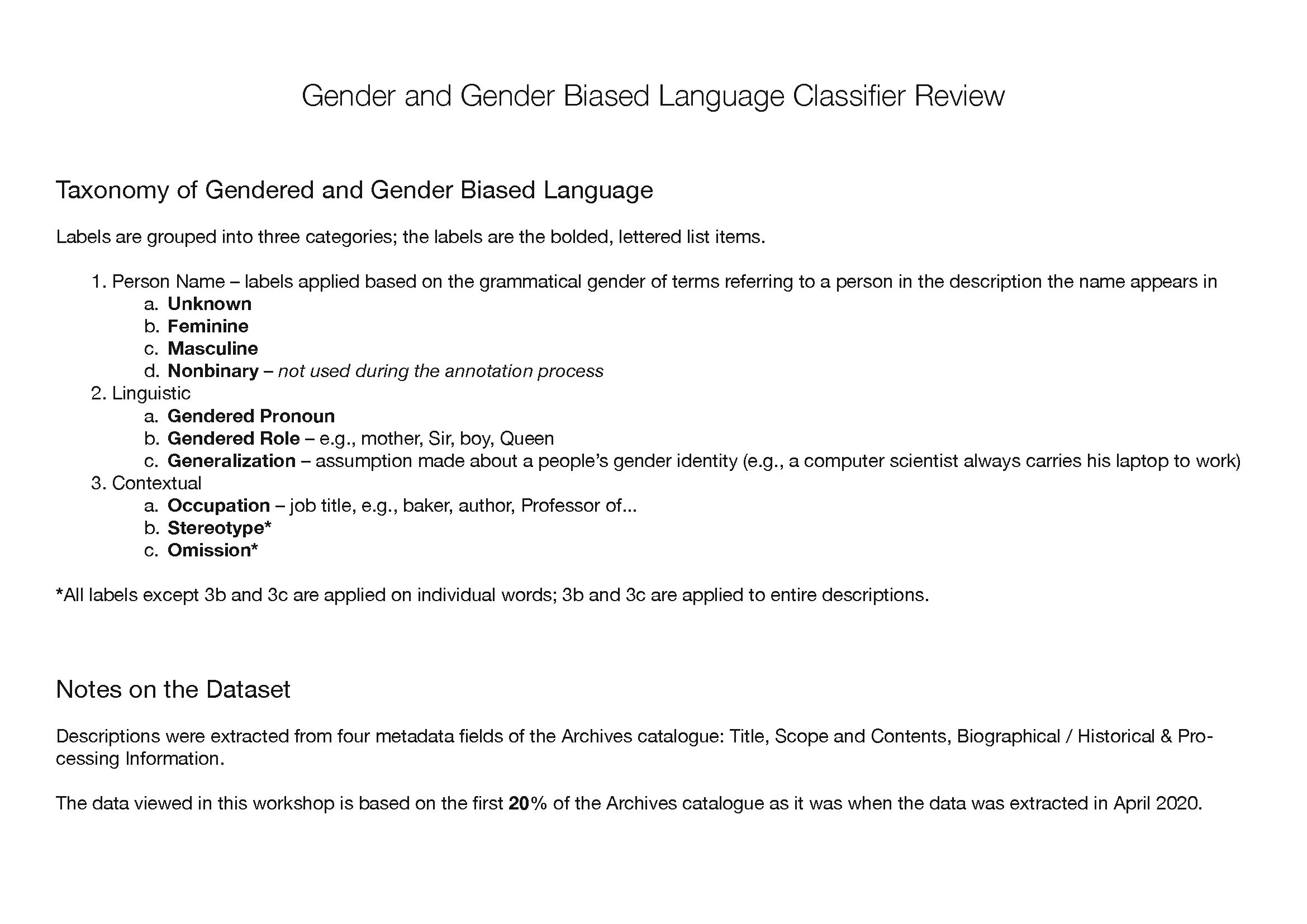}
        \caption{The front of the worksheet participants were given for the workshop's first activity displaying the coding taxonomy that outlines how the ML models were trained to classify descriptions.}\label{fig:worksheet1-1}
        \Description{The figure displays the front of a worksheet distributed during the workshop's first activity.  The worksheet lists the codes that ML models were trained to classify text using the human-coded data from Havens \textit{et al.} (2022) as the the training dataset.  Below the codes list, the worksheet explains that the data visualized on the workshop's two worksheets represent the first 20\% of an archival catalog's metadata descriptions as of April 2020.}
        \centering
        \includegraphics[width=0.75\linewidth]{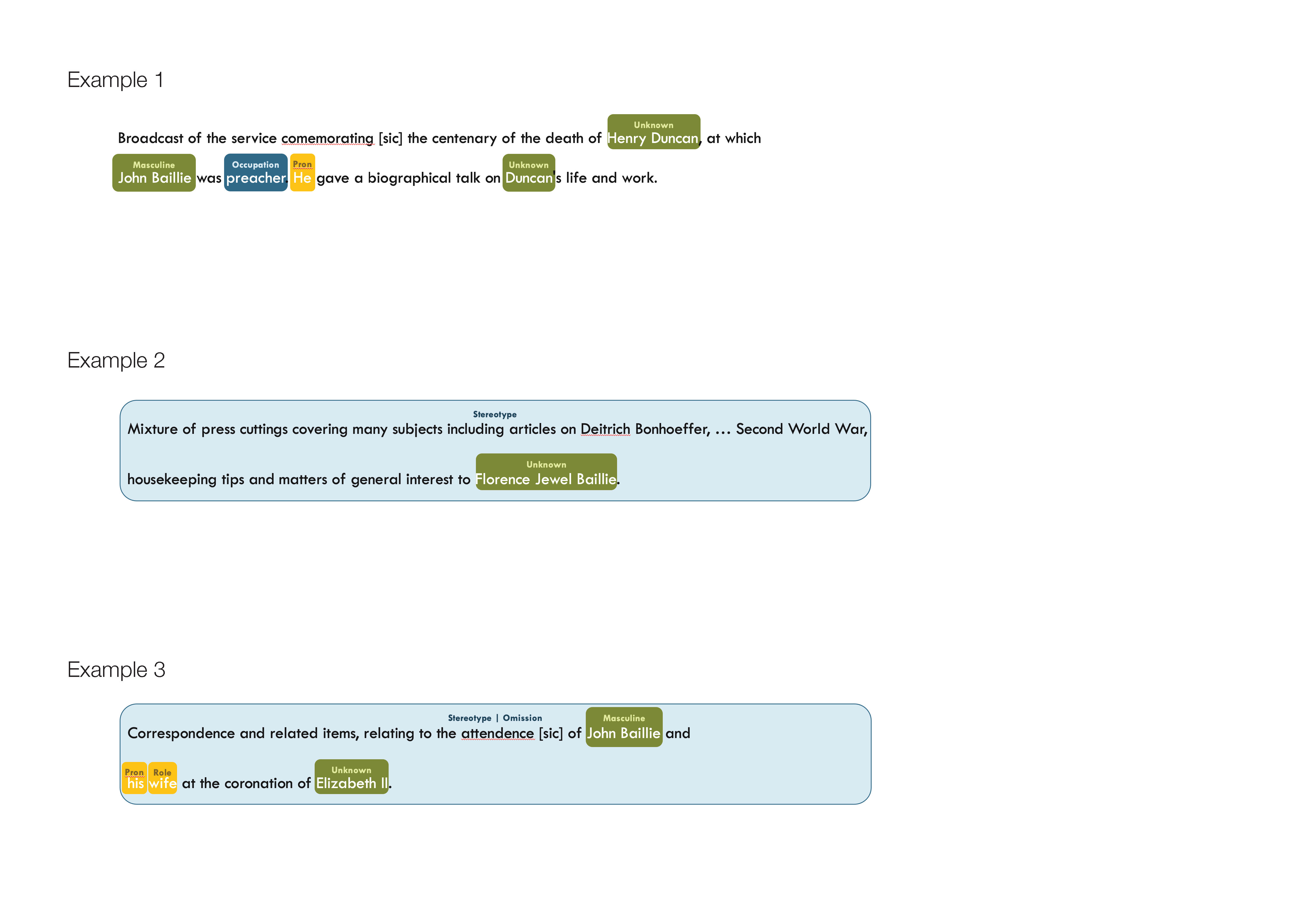}
        \caption{The back of the worksheet participants were given for the workshop's first activity with three metadata descriptions from the University of Edinburgh's archival catalog, coded according to the taxonomy (Table \ref{t:taxonomy}).  Together the three descriptions provide examples of how to apply every code in the taxonomy.}\label{fig:worksheet1-2}
        \Description{The figure displays the back of the first workshop activity's worksheet, which has three GLAM catalog descriptions overlaid with rectangles to indicate the coding of text.  This side of the worksheet has notes written in pencil by a workshop participant that comment on the descriptions and their coding.}
    \end{figure*}

    The workshop consisted of an introduction, two activities, and a wrap-up question, lasting two hours total.  We began the workshop by explaining the supervised, traditional ML method we used to create text classification models, which we had used to classify descriptions from the University of Edinburgh's archival catalog (\textit{i.e.}, the participants' GLAM institution's catalog) for gender bias.  Then we began Activity 1 (A1), which focused on investigating RQ1: \textit{How do information and heritage professionals in the GLAM sector conceptualize bias?  To what extent does this align with how ML researchers and practitioners conceptualize bias?}  We gave each participant a worksheet with the codes that the models were trained to classify with on one side (Figure \ref{fig:worksheet1-1}) and three coded descriptions on the other side (Figure \ref{fig:worksheet1-2}).  We selected the coded descriptions from our models' training data because we aimed the discussion during A1 to focus on the interpretation of the descriptions and meaning of the codes, facilitating a human-centered evaluation of the models' design.  To start the discussion, we asked participants:
    \begin{itemize}
        \item Q1.1: Do you agree or disagree with, or are you unsure about, the labels on the descriptions?  Why?
        \item Q1.2: How would you use this information?
        \item Q1.3 What information is missing that you would need to support your work?
    \end{itemize}
    A1 lasted one hour.  We took a 10-minute break and then began the second activity.
  
\begin{figure*}
        \centering
        \includegraphics[width=0.7\linewidth]{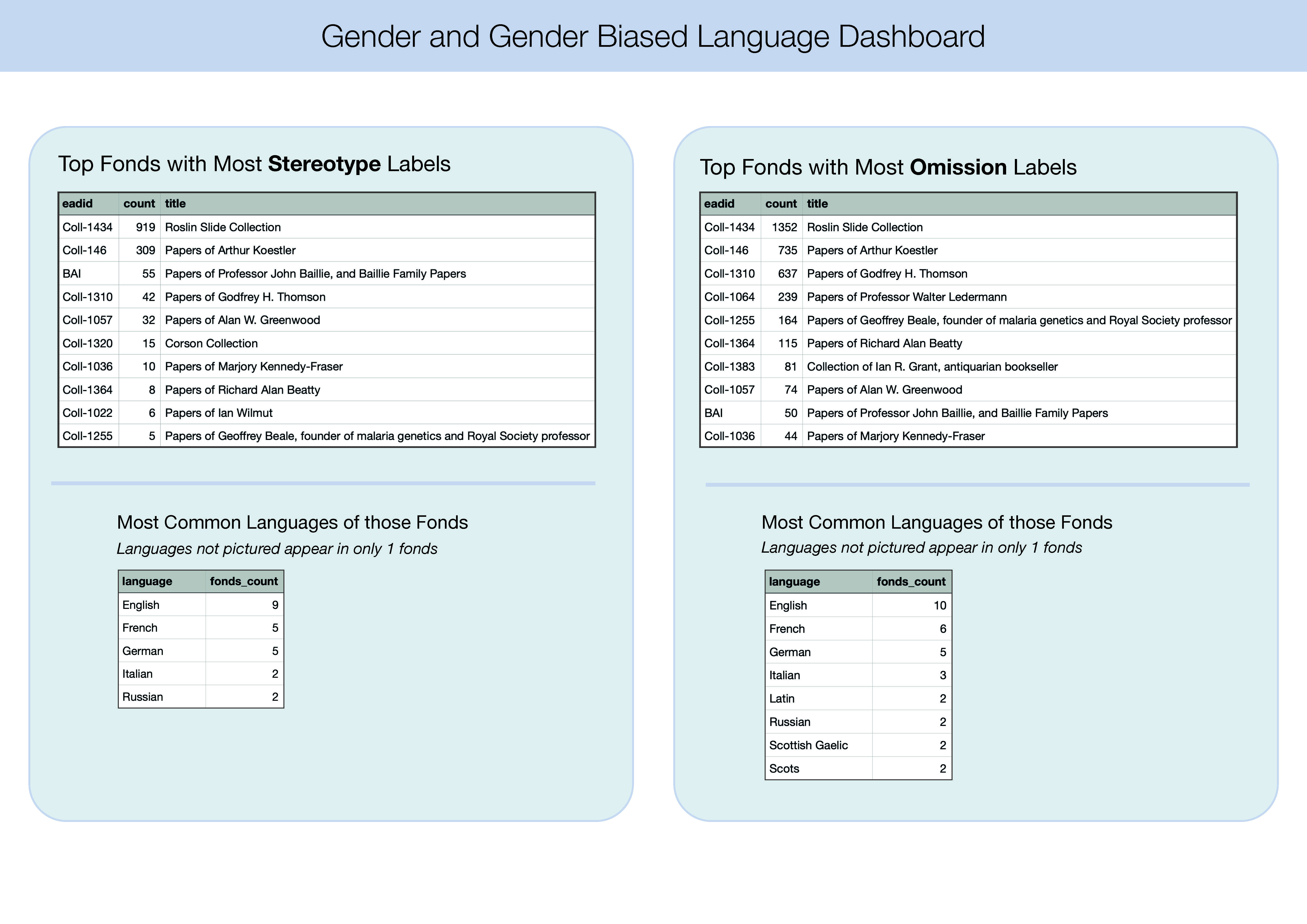}
        \caption{The front of of the worksheet participants were given for the workshop's second activity. The tables at the top show total quantities of \textbf{\textit{Stereotype}} and \textbf{\textit{Omission}} codes in the collections (``fonds'') with the highest total counts of those codes.  The tables at the bottom show the most common languages of the archival records in the collections listed in the tables above.}\label{fig:worksheet2-1}
        \Description{This figure displays the front of the second workshop activity's worksheet, which is titled, ``Gender and Gender Biased Language Dashboard.''  The top left of the worksheet has a table listing the top 10 collections that the ML models gave with the most Stereotype codes, ranging from 919 to 5.  The bottom left has a table that lists the most common language of the material in those collections, which in descending order are English, French, German, Italian, and Russian.  The top right has a table listing the top 10 collections that the models gave with the most Omission codes, ranging from 1,352 to 44.  The bottom right has a table listing the most common language of the material in those collections, which in descending order are English, French, German, Italian, Latin, Russian, Scottish Gaelic, and Scots.  The worksheet has a few notes written at the top and bottom of the page by a workshop participant.}
    \end{figure*}

\begin{figure*}[ht!]
        \centering
        \includegraphics[width=0.7\linewidth]{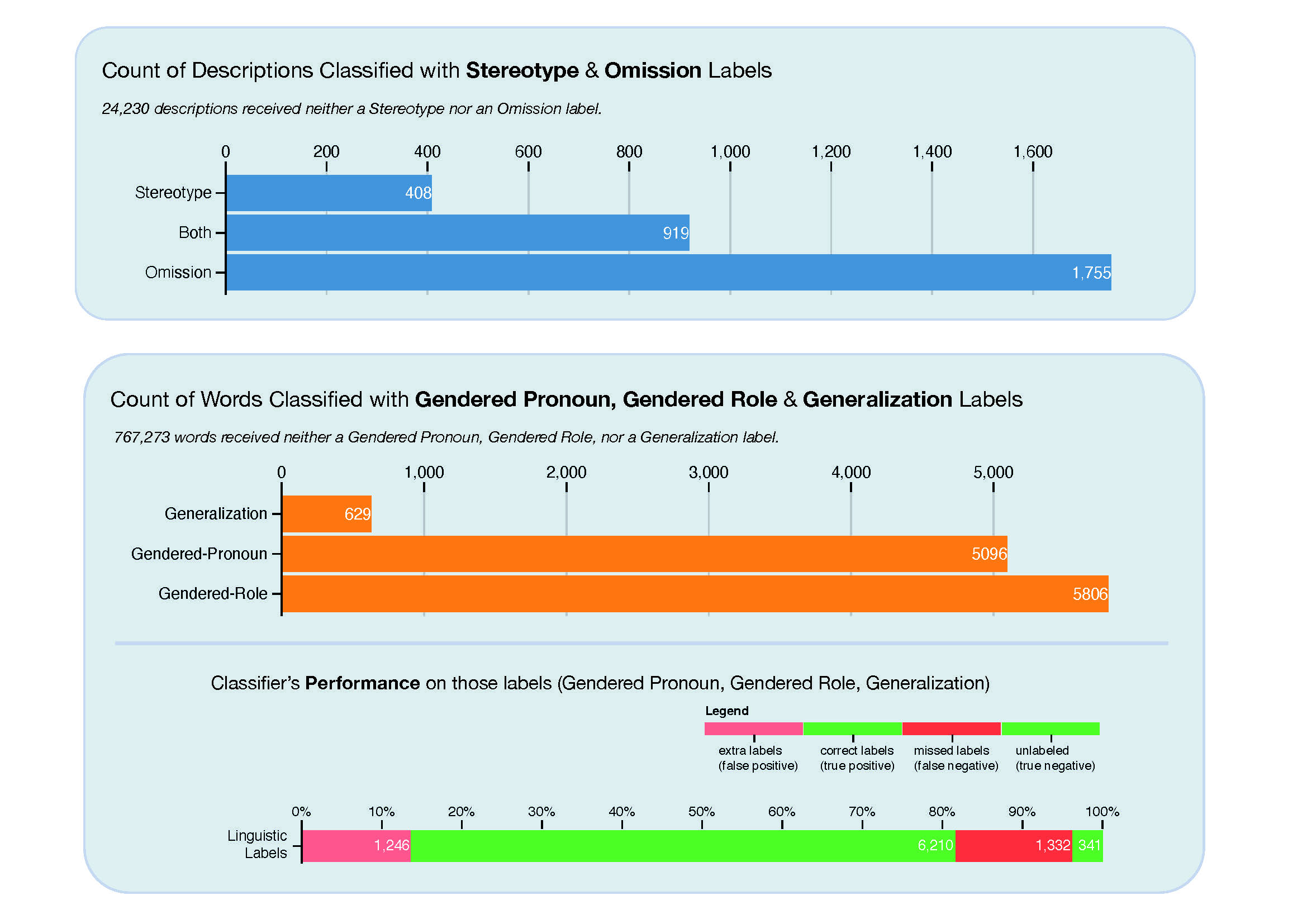}
        \caption{The back of of the worksheet participants were given for the workshop's second activity displaying two charts with quantities of taxonomy codes across the entire training dataset, and one chart visualizing the performance of a classifier as represented by true positive, false positive, true negative, and false negative counts.}\label{fig:worksheet2-2}
        \Description{This figure displays three charts about the ML models' classification of GLAM catalog descriptions.  At the top, a horizontal bar chart displaying, from top to bottom in ascending order, the total counts of descriptions classified with either only a Stereotype code, an Omission and a Stereotype code, or only an Omission code.  The middle chart is a horizontal bar chart displaying the count of words the ML model classified with, from top to bottom in ascending order, the Generalization code, the Gendered Pronoun code, and the Gendered Role code.  The bottom chart displays the ML model's performance classifying descriptions' text with these codes, showing the quantities and percentages of extra codes (false positives), correct codes (true positives), missed codes (false negatives), and correctly not coded (true negatives).  There are brief notes written by a participant in pencil by the top and middle charts.}
    \end{figure*}

    Activity 2 (A2) focused on RQ2: \textit{How can the identification of gender biased language with ML models inform and impact information and heritage professionals' understanding of gender biased language and management of bias in their institutions' cultural heritage collections?}  We gave each participant a worksheet (figures \ref{fig:worksheet2-1} and \ref{fig:worksheet2-2}) with tables and charts displaying quantities of gendered and gender biased language in the University of Edinburgh's archival catalog, and a chart displaying performance measures for the Linguistic Classifier.  We calculated these quantities based on the output codes of Cascade 2's LC and OSC and the output codes of the baseline PNOC because these were the best-performing versions of our models (see \ref{a:ml-results}).  We presented participants with these quantities to facilitate a human-centered evaluation of our models' outputs and performance.  To start the discussion, we asked participants:
    \begin{itemize}
        \item Q2.1: What do you understand from the information on the worksheet?  What questions do you have?
        \item Q2.2: How would you use this information?
        \item Q2.3: What information is missing that you would need to support your work?
    \end{itemize}
    A2 lasted 40 minutes.

    We concluded the workshop by asking participants one wrap-up question:
    \begin{itemize}
        \item Q3.1: Is there any information on the worksheets you would want to share with visitors (\textit{i.e.}, researchers, the general public)?
    \end{itemize}
    The wrap-up activity lasted five minutes.  We sent participants a questionnaire consisting of 18 multiple choice and short answer questions after the workshop.

    \subsection{Participants}\label{participants}
    
    We recruited participants through an email to 14 information and heritage professionals working on the University of Edinburgh's Heritage Collections teams, 10 of whom participated.  According to our post-workshop questionnaire, participants' job titles can be broadly categorized as archivist (six participants), curator (one participant), librarian (one participant), and manager (one participant).  They described their job responsibilities as including acquisition, cataloging, curation, processing, development, management, preservation, interpretation, and facilitation of discovery and access of collections; teaching and giving training about collections; overseeing colleagues and volunteers; responding to inquiries about collections; communicating with donors and stakeholders of collections; participating in international communities of practice; and research.  They reported encountering biases in GLAM catalogs and collections related to gender, sexuality, faith, race and ethnicity, nationality, language, accent, and disability.  Participants' GLAM work experience ranged from six years to over 10 years.  They described their gender as ``female,'' ``cisgender female,'' and ``female-ish.''


    \subsection{Analysis}
    
    We performed content analysis on the workshop discussion using a grounded theory approach~\cite{Glaser_Strauss_1980,Krippendorff_2018,Robson_McCartan_2016}, performing open coding on the transcript of the discussion.  Our approach to open coding was rooted in intersectional feminism~\cite{Crenshaw_1989,HillCollins_2000} and critical discourse analysis~\cite{Fairclough_2003,Bucholtz_2003}.  We also referenced participants' responses to the post-workshop questionnaire and their notes on the worksheets during our content analysis.  We report the results of this analysis as ``observations'' (\S\ref{observations}), which we shared with workshop participants prior to publishing to provide an opportunity for clarification and feedback.  From these observations we identified three areas of value the models held for information and heritage professionals' management of bias, and three characteristics of data bias for ML researchers and practitioners' consideration (\S\ref{discussion}).

\section{Results}\label{observations}
Below we summarize 15 observations (O1-O15) that emerged from our content analysis, grouping the observations by our research questions. Participants are referred to anonymously as P1, P2, \textit{etc.}

    \subsection{RQ1: Conceptualizations of Bias}\label{observations-rq1}
    Regarding observations relevant to RQ1, participants' comments communicated a conceptualization of bias that is context-dependent, changing based on who is represented in the data and who is looking at the data, including when and where those people are located and those people's lived experiences.
    
        \subsubsection*{O1: Concern with losing gender information.} 
        At the beginning of the workshop, participants' discussion communicated a concern about the codes highlighting language that should be removed or changed.  P6 explained how, due to the historical and ongoing marginalization of women in many societies around the world, often a woman is only identified in cultural heritage records with a title and last name (\textit{e.g.}, ``Mrs. MacDonald''), so, \textit{``if you lose `Mrs.' then you lose all of their identity rather than enhancing their identity.''}  P8 gave an example of how the identity of a woman, who was depicted in a painting with her husband, was able to be discovered thanks to the more extensive documentation of her husband and the description of the woman as a ``Mrs.''  Through discovering the woman's identity, P8 was able to learn that the wife brought significant funds to she and her husband's marriage that enabled the couple to develop a substantial art collection.  The inclusion of grammatically and lexically gendered terms can thus provide clues to guide research into the contributions of women, supporting efforts to address the lack of documentation about women's roles and contributions throughout history~\cite{Beard_2017,Graeber_Wengrow_2021,Hessel_2023}.

        Participants also discussed the value of identifying people by gender even if that gender is assumed.  Regarding the application of the \textbf{\textit{Unknown}} code, P7 stated, 
        \textit{``I feel kind of weird, though, with it, in that now I can't say what their genders are [...] how can we say who they are if we can't assume that?''}
        Participants explained how describing people with gendered terminology can be useful even if an assumption was made about a person's gender, because that assumption reflects how the person was perceived in society during their lifetime.  During the workshop, in response to participants' concern about losing information with the removal of gendered terminology, we clarified that the taxonomy's codes were not intended to highlight language for removal or change, but rather were intended to make potential gender biases more easily identifiable.
    
        \subsubsection*{O2: Gender bias in the codes' application.} 
        Regarding the application of the \textbf{\textit{Occupation}} code, participants suggested that it could have been applied more broadly than only to job titles.  Looking at examples 2 and 3 in Figure \ref{fig:worksheet1-2}, P10 asked why ``housekeeping'' did not have an \textbf{\textit{Occupation}} code and then said, \textit{``And it's the same thing, you know, with `coronation,' hinting at her [Queen Elizabeth II's] occupation.}  Despite the intention of the codes to push back against such gender biases~\cite{Havens_2022}, their application reinforced occupational gender biases and the under-documentation of women's historical contributions.  Women have faced greater barriers to entering the workforce and obtaining job titles, but this does not mean women have not worked.

        \subsubsection*{O3: Subjectivity of bias.} 
        Differences  in the interpretations of the codes, as well as an awareness of the subjectivity of bias and language, came through in participants' comments and questions.  \textbf{\textit{Stereotype}} yielded the greatest debate.  Participants asked questions about how \textbf{\textit{Stereotype}} had been defined and requested additional examples of text that would be classified with this code.  P5 spoke of the contextual nature of a stereotype, stating, \textit{``It's changed by people, by culture, by things like this, so it seems really hard to define what it is.''}  P7 expressed concern about the judgment implied with the \textbf{\textit{Stereotype}} code in the second description (Figure \ref{fig:worksheet1-2}), stating, \textit{``You could be doing her a huge injustice if her life's mission was housekeeping.''}  This description prompted the greatest discussion about stereotypes.  P8 requested clarification about whether the \textbf{\textit{Stereotype}} code was in fact applied due to the association of a woman with ``housekeeping.''  P3 responded:
        \begin{quote}
            \textit{``I thought it wasn't just the housekeeping but that a woman's interests just get summarized into `general interests.' [...] It doesn't just say, `Florence Jewel Baillie was an avid researcher of Dietrich Bonhoeffer, the Second World War, housekeeping,} etc.\textit{'  It just says here are some news cuttings on some things that she read about.  Which kind of makes a judgment.  That's what I thought the stereotype was in this instance.''}
        \end{quote}

        \subsubsection*{O4: Value of dates to contextualize data.} 
        Participants expressed the importance of including dates where available because dates situate a description's language in the historical context in which the language was produced.  Regarding potential biases of descriptions in the ``Title'' catalog metadata field, P3 said, \textit{``we probably inherit that [...] so it'd be good to be able to isolate language that's been created recently by an archivist.''}  P1 and P3 spoke of the value of dates alongside a cataloging system that would maintain the version history of a description, which is not provided in current cataloging systems' functionality.  P1 explained, \textit{``it's about dating descriptions and not overwriting them.''}

        \subsubsection*{O5: Sense of responsibility to past, present, and future stakeholders of cataloged cultural heritage.} 
        Participants communicated a sense of responsibility to historical people and communities represented in cultural heritage records, to present-day descendants of those people and members of those communities, and to future visitors who will search GLAM catalogs for records of those people and communities.  Reflecting on the role of GLAM catalogs and GLAM description practices, P9 commented:
        \begin{quote}
            \textit{``I think it's sort of socially really important [...] because so many people are going to look it [the catalog] up, students or members of the public, and refer to that as a kind of non-biased factual resource.  So it's different from writing something where you assess your positionality and make that clear to the reader [...] it's a big responsibility to get right as well.''}
        \end{quote}
        
        Participants expressed concern about the challenge of navigating this responsibility when different stakeholders hold conflicting preferences or points of view.  Recalling cataloging work with cultural heritage records in which a person referred to themselves as a ``Highland traveler,'' a term today considered derogatory, P6 recounted how a present-day member of this community had contacted the University of Edinburgh's Archive to request that the catalog description be changed to use today's accepted term, ``Gypsy Roma Traveler.'' P6 explained the conflicting responsibilities information and heritage professionals must grapple with in such situations:
        \textit{``So you then get into layers of, well who's say is it?  Is it the current community or is it the person whose own voice it is?''}
        P4 gave an example of this conflict regarding gender, explaining that historical members of the trans community used terms to describe themselves that are considered offensive today.  Nonetheless, P4 viewed these terms as important to include in catalog descriptions of cultural heritage records because such terminology is, \textit{``something that's an example of trans history or gender non-conformity.''}
        
        In response to this discussion, P7 asked, \textit{``What do you do when [...] in 20 years' time, something you created is deemed offensive?''}  P2 added that diachronic changes in language are unpredictable, with certain terminology evolving from having positive or neutral to offensive connotations (\textit{e.g.}, ``mistress'') while other terminology evolves from having offensive to positive or neutral connotations (\textit{e.g.}, ``queer'')~\cite{Schulz_1975,Shopland_2020}.  Participants came to a consensus that the best approach to this challenge is to include as much terminology as possible with contextualization, though they also expressed dissatisfaction with how this approach could make descriptions \textit{``clunky-looking''} or \textit{``difficult to read.''}

        \subsubsection*{O6: Barriers posed by institutional and professional structures.} 
        While acknowledging the power held as the people describing collections, participants also discussed limitations to their power.  Regarding institutional barriers faced by information and heritage professionals when requesting cataloging resources, P1 stated, \textit{``We will always get back to needing an army of catalogers,''} referring to the impossibility of ever hiring enough people to describe a GLAM institution's cultural heritage records to a satisfactory level of detail.  Regarding the limited power that information and heritage professionals have over GLAM cataloging standards and expectations for interoperable GLAM systems, participants discussed a tension between standards and  the reality of cultural heritage records.  P3 stated that social biases are encoded not only in cataloging standards and systems built to use those standards, but also \textit{``the whole profession.''}  P1 asked rhetorically, \textit{``Who's on the boards that get to deem that this [}e.g.\textit{, the standards' terminology] is the right language?''}  Implicit in participants' discussion was the assumption that social biases are inevitable.  Nonetheless, they spoke of ways to use the tools of their profession to contribute to efforts that empower, rather than further oppress, marginalized communities of people.  P1 stated, \textit{``We were taught to accept them [the standards] and that's, well, I don't want to.  I want to see them as a framework.''}

        \subsubsection*{O7: Desire for more time to document collections.} 
        Participants spoke of time constraints that limit the research they can conduct on cultural heritage records to inform their descriptions of the records in catalogs.  P3 spoke of the implications of digital technologies, stating, \textit{``For digital archives, at the scale some of them are, the human race will die out before we'd be able to catalog every website.''}  P8 spoke of the mismatch between the time needed and the time available for describing cultural heritage records that the participant was currently cataloging: \textit{``[...] then you have to research her relationship, his relationship to get a hint as to what Mary's role was as well, and that takes you into a three year Ph.D.!  And you've got six months to catalog the collection!''}  Participants saw an unresolvable tension between making as many records as possible discoverable (\textit{i.e.}, writing descriptions that researchers, students, and the general public can query in online catalogs) and conducting lengthier research into cultural heritage records that could challenge the social biases they may reflect.

    \subsection{RQ2: ML for Informing Understandings and Management of Bias}
    Regarding observations relevant to RQ2, participants' comments communicated an interest in integrating ML into existing workflows but not in fully automating description writing or description review processes.

        \subsubsection*{O8: Value of gendered language codes individually vs. in combination for understanding collections' biases.} 
        Participants expressed agreement regarding the application of the \textbf{\textit{Gendered Role}} code.  P8 communicated an interest in seeing the quantity of different text spans coded with \textbf{\textit{Gendered Role}}, stating that those quantities could support the search for \textit{``bias and gaps''} in a catalog's descriptions.  P8 explained that this code would be especially interesting for identifying \textit{``where a `Mrs.' is, and saying that's potentially a partner that needs further exploration,''} referring to gaps in historical documentation about women's work and contributions~\cite{Beard_2017,Graeber_Wengrow_2021,Hessel_2023}.

        Participants expressed disagreement regarding the application of the \textbf{\textit{Gendered Pronoun}} and \textbf{\textit{Unknown}} codes.  P2 and P5 expressed interest in seeing gendered pronouns annotated if there was an assumption being made about a gender group implicitly communicated through the use of the pronoun (\textit{i.e.}, the language that the \textbf{\textit{Generalization}} code is intended to annotate).  These participants were not interested in seeing the code applied to every instance of a gendered pronoun across a catalog's descriptions.  P7 expressed disagreement with the \textbf{\textit{Unknown}} code's application to one instance of a person's name being based on grammatically or lexically gendered terminology provided only in the description in which that instance of the name appeared, rather than across all descriptions in which that person's name appeared.  P10 communicated a similar disagreement, pointing to the third description (Figure \ref{fig:worksheet1-2}) in which ``Elizabeth II'' had been coded as \textbf{\textit{Unknown}}: \textit{``In this case it's a woman in power, but then we're not recognizing that there were some women in power, if everybody's not identified by gender.''}  This disagreement relates to participants' sentiments about the value of gendered language as summarized in O1.  
        
        Conversely, P1 expressed agreement with the codes' application when considered in combination.  P1 stated, \textit{``It's not about just the} \textbf{Gendered Pronoun}\textit{, it's actually when they come together, they produce something that's really useful.''}  This participant felt that the codes as a whole contributed to a better understanding of the gender biases in cultural heritage records and descriptions of them in catalogs, because together the codes provide a large-scale interpretation of a catalog's descriptive language.  P1 viewed the ML models and their classification of descriptions with these codes as tools to incorporate into information and heritage professionals' workflows, with the information and heritage professionals determining when and how to apply the tools.

        \subsubsection*{O9: Cautious interest in using ML to support investigations into collections' biases.} 
        Participants communicated both skepticism and interest in the ML models' ability to inform their understanding of gender biases in collections and collections' descriptions.  P1 proposed, \textit{``I wonder whether actually, if there were collections we would suggest putting through the models to test them, to see if they pick up on things that we know are in those collections, whether that would be something useful.''}  Other participants confirmed that they would be interested in reviewing model-made codes on descriptions of cultural heritage collections that they were familiar with to evaluate the ML models' relevance to their understanding of collections' gender biases.  That being said, some participants saw the ML models' classifications of potentially gender biased language useful as an initial step for critical cataloging workflows.  Regarding applications of the models, P9 said, \textit{``it would maybe enable this kind of revisiting and correcting to be done more quickly and with less labor.''}  Critical cataloging could be a use case for human-ML collaboration, with the ML models selecting a subset of a catalog's descriptions for an information or heritage professional to review, rather than the information or heritage professional having to start with the entire catalog.

        \subsubsection*{O10: Models as a tool to integrate into existing workflows to improve awareness of collections' biases.} 
        Participants explained that upon being presented with ML models' classifications of gender biased language in descriptions, the next step they would want to take would be to have an information or heritage professional manually review the classified descriptions.  P5 viewed the models as, \textit{``a way to flag something that might be problematic.  But then you've got to have an archivist or cataloger go back,''} rather than using the models as the final decision-maker on whether additions or changes should be made to a description.  Participants also saw the models' classifications as useful for informing guidelines for describing cultural heritage records, particularly for students and volunteers who are writing descriptions.  P2 proposed that the models' output could inform a \textit{``how-to guide''} for describing collection material, to which P8 responded, \textit{``I think that would be quite useful to have,''} explaining how such a guide would be helpful for that participant's current work with volunteer catalogers.

        \subsubsection*{O11: Models as a tool for self-reflection on biases encoded in one's own language choices for descriptions.} 
        Participants also discussed the value of the models for facilitating their own reflection on how they describe cultural heritage collections.  Participants spoke of how concepts of bias evolve and how different stakeholders of a collection may have different opinions about which terminology is correct and which is biased, as further described in O5.  Rather than trying to fix or remove bias, P1 proposed that the aim could be awareness, with the ML models helping participants in \textit{``being aware of our practice, or aware of where we've taken old catalogs and perpetuated things.''}\footnote{When a GLAM institution acquires a new collection of cultural heritage records, depending on the resources the institution has available, information and heritage professionals may quote from descriptions the collector (\textit{i.e.}, the person or people from whom the institution acquired the collection) already wrote of the cultural heritage records in the institution's catalog, because GLAM institutions' primary aim is to make records discoverable as quickly as possible.}  

        \subsubsection*{O12: Model outputs as evidence to support resource requests.} 
        P1 spoke of how the models could provide evidence of the need for particular description and cataloging projects.  P1 discussed the models' classifications as \textit{``an evidence base''} to support requests for hiring interns and full-time information and heritage professionals.  This participant saw the models' coding of descriptions as a way to demonstrate that, \textit{``there is a skill to this [description] and these skills of archivists, librarians, and curators are needed in this space.''} 

        \subsubsection*{O13: Uncertainty about ML models' capabilities.} 
        Upon reviewing the tables of collections with the highest counts of descriptions that were model-coded with \textbf{\textit{Omission}} or \textbf{\textit{Stereotype}}, P1 expressed interest in three of the collections that worksheet 2 listed (Figure \ref{fig:worksheet2-1}).  Those collections' presence in the tables aligned with P1's knowledge of them as containing records regarding eugenics, misogyny, imperialism, and gender stereotyping.  P1 stated, \textit{``I'm absolutely fascinated that Godfrey Thomson\footnote{\url{archives.collections.ed.ac.uk/repositories/2/resources/85818}} has come up,''} explaining that the high \textbf{\textit{Stereotype}} counts could be due to \textit{``the time period he worked in but also the environment.  I think some of his work early on looked at eugenics and that kind of classification of people.''}  That being said, P1 expressed uncertainty about whether the models had labeled these collections for the reasons she expected, stating, \textit{``we'd have to investigate further.''}

        P7's comments expressed the strongest doubt in the models' capabilities relative to those of the other participants.  In regards to the presence of the \textit{Roslin Slide Collection}\footnote{\url{archives.collections.ed.ac.uk/repositories/2/resources/85706}} in the tables recording collections with high \textbf{\textit{Omission}} and \textbf{\textit{Stereotype}} counts, P7 stated, \textit{``I suspect there is some complicated problem with the data and the collection, and the way the data is being compiled, which is why it's got such high numbers there.''}  P1 responded that the presence of the \textit{Roslin} collection in the tables did not surprise her because \textit{``The images [in the collection] were collected during the height of empire and colonialism, and there is language within it that is gendered.  There's also how women are represented and that can be in terms of different parts of the world and communities.''}

        \subsubsection*{O14: Little interest in quantitative model evaluation measures.} 
        Participants did not ask questions or comment on the second worksheet's charts about the models' performance measured with metrics that inform standard ML evaluations, namely true positives, false positives, and false negatives (Figure \ref{fig:worksheet2-2}).  Handwritten notes on participants' worksheets suggest that participants were more interested in gaining an understanding of the type of language that the models classified with the taxonomy's codes.

        \subsubsection*{O15: Distinct needs of GLAM employees and visitors.} 
        In response to our wrap up question about sharing information on the worksheets that the ML models provided with visitors to the collections (Q3.1), P1 stated, \textit{``I think it'd be really good to ask some of our User Services people who get inquiries in.''}  Another participant stated, \textit{``It'd be useful to speak to users [...] because I've got a feeling they come with knowledge that we don't have,''} recalling discussion earlier in the workshop about descriptions of cultural heritage records being works-in-progress intended to help researchers discover potential historical gaps they can investigate further.  The results of such an investigation could be brought back to the information professionals for \textit{``enhancing or building upon''} a catalog's existing descriptions.

        During the workshop's pilot, a participant's comment on the second worksheet's charts of cross-collection measurements (Figure \ref{fig:worksheet2-2}) provide some insight on information a visitor to a catalog might benefit from seeing.  The participant stated, \textit{``I just don't really think about the fact that they [GLAM catalog metadata descriptions] are created in a certain time period by certain people.  So I think it adds a layer of something to think about when you're looking for materials [...] it's a reminder as a researcher that this is not some sort of neutral, objective tag of what you can actually find.''} 

\section{Discussion}\label{discussion}
Predominant approaches to bias in ML aim to measure and remove bias, yet they have not adequately studied how bias may manifest in data.  As a result, the inequitable power relationships in society that cause bias in ML systems are rarely identified, let alone challenged, with these approaches.  Conversely, the human-centered ML approach we implemented reframes the problem of ML bias, aiming to identify gender bias so it can be communicated.  Making bias visible in this way informs and motivates efforts to challenge cisnormative, patriarchal societal structures that create and reinforce gender bias, such as Hessel's narrative of art history that centers women~\cite{Hessel_2023} and the Trans Metadata Collective's cataloging resource, \textit{Best Practices for Describing Trans and Gender Diverse Resources}~\cite{TMDC}.  Our work aligns with D'Ignazio's ``restorative/transformative data science,'' which leverages data science methods to analyze power, reframe problems, challenge hegemonic worldviews, and produce data~\cite{DIgnazio_2024}.  By investigating how bias may manifest in data and the limitations of ML methods for identifying those manifestations, we contribute evidence in support of calls to expand ML bias and fairness approaches~\cite{Birhane_2020,Blodgett_2020,CorbettDavies_2023,Klein_D’Ignazio_2024,Goree_Crandall_2023}.  Uniquely, we have offered an approach for determining the feasibility of removing bias or achieving fairness in a given ML use case.

    \subsection{RQ1: Conceptualizations of Bias}
    
    Participants approached our ML models with caution due to concerns about the contextual information that ML models cannot reliably encode.  They reject the assumption that computational technologies alone always offer the best solution or approach, a form of bias that Broussard terms ``technochauvinism''~\cite{Broussard_2023}.  Instead, participants viewed the models as valuable for a ``human-in-the-loop'' system, where ``a machine does a lot of the work but meaningful human work and meaningful human interaction are prioritized'' (\textit{ibid.}).  This definition of \textit{human-in-the-loop} differs from how such systems are described in ML literature, which tends to focus on using human feedback to augment data and manipulate models, integrating human labor into a predominantly ML-driven process~\cite{Wang_2021}.  
    
    From the 15 observations reported in \S\ref{observations}, we identified three characteristics of bias as conceptualized by information and heritage professionals:
    \begin{enumerate}
        \item Data bias is contextual, meaning it is a result of complex interactions between cultural, historical, economic, political, and social events, relationships, and practices that are constantly evolving (O3, O4, O5, O6, O15).
        
        \item The values or needs of different communities of people may conflict, so the approach to managing data bias that one community prefers may be considered harmful by another community (O1, O5).
        
        \item Data bias is inevitable, because the choice of what data to collect and not collect, and how to interpret the collected data, are processes that privilege certain perspectives at the expense of others (O2, O3, O11).
    \end{enumerate}
    
    These characteristics motivate information and heritage professionals' desire to use ML to enhance, rather than fully automate, critical cataloging.  Furthermore, they illustrate how information and heritage professionals' conceptualization of bias, which is based on the external relations of data, differs from predominant conceptualizations of bias among ML researchers and practitioners.  Although there are exceptions (\textit{e.g.},~\cite{Beelen_2022}), most ML approaches to bias are based on the internal relations of data, meaning characteristics that are mathematically quantifiable in word embeddings or comparisons of a model's performance for different demographic groups (\S\ref{related-work}).  Consequently, the ML community continues to deploy models without adequately accounting for the contextual nature of bias, and the models' widespread adoption harms already-marginalized communities~\cite{Hofmann_2024}.  Though existing research has considered values in ML systems~\cite{Birhane_2022,Kasirzadeh_Gabriel_2023}, more work is needed investigating value trade-offs, considering how prioritizing one community's values can harm another community.  Additionally, while removing bias and aiming to create fair ML systems is suitable for certain use cases (\textit{e.g.},~\cite{Cao_2020}), our workshop findings demonstrate that ML approaches to bias must also consider how to manage bias when it cannot or should not be removed.
    
    Notably, participants never spoke of removing bias or neutralizing the language of a description.  Instead, participants spoke of how a person's understanding of what is biased changes over time and varies depending on the relationship between people who create, are represented in, interpret, and use the data.  This aligns with feminist understandings of data and knowledge as \textit{situated}~\cite{Harding_1995,Haraway_1988,HillCollins_2000} and \textit{intersectional}~\cite{Crenshaw_1989,D_K_2020}, where people's experiences of privilege and oppression result from the interaction of multiple identity characteristics within particular cultural, historical, economic, political, social, and temporal contexts.  For GLAM workflows, choosing what to describe in detail and what to summarize more vaguely inevitably privileges certain points of view over others~\cite{Duff_Harris_2002}.
    
    Through first-hand experience creating data to describe collections, participants were well aware that managing data biases will always be an imperfect, work-in-progress effort with no universally-applicable solution~\cite{Caswell_Cifor_2016,Tai_2021,Bowker_Star_2000}.  Participants acknowledged the power they hold in writing descriptions of heritage collections that people will reference \textit{``as a kind of non-biased factual resource''} (O5), as well as limitations on that power from societal, institutional, and professional structures (O5, O6).  
\begin{figure*}[t]
    \centering
    \includegraphics[width=0.8\linewidth]{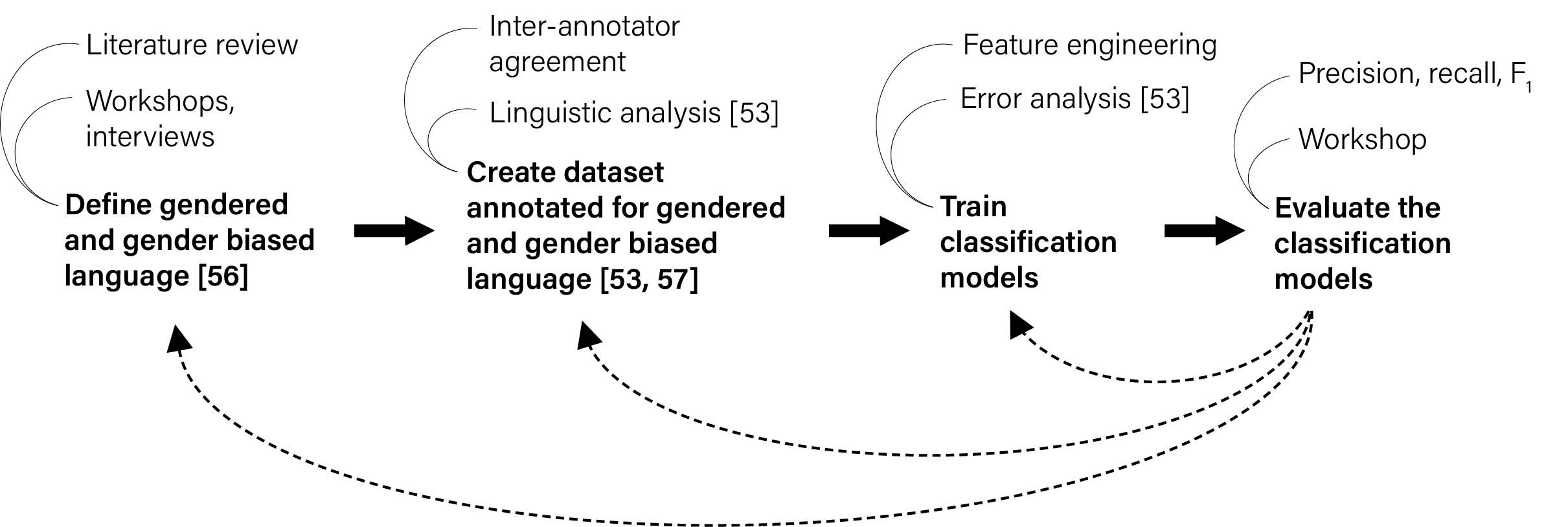}
    \caption{The same flow chart of our four-step research process as Figure \ref{fig:research-process} with three dotted line arrows added.  These new arrows connect the evaluation step to each of the previous steps, indicating the iterations that future work can follow to refine how a model's task is defined or conceptualized, refine the contents and quality of a dataset, and refine the design of a model.  Using traditional ML approaches enables these iterations to be undertaken quickly and at low cost.  Once the quantitative and qualitative evaluations of a model are satisfactory, deep learning approaches could be used to create a model on the iteratively refined training dataset.}
    \label{fig:future-process}
    \Description{The same flow chart as figure 1, depicting our four-step research process, with three new dotted line arrows added to indicate optional iterations back to previous steps from the final, evaluation step.  From left to right, the steps are: (1) define gendered and gender biased language, (2) create dataset annotated for gendered and gender biased language, (3) train classification models, and (4) evaluate the classification models.  Radiating out and above the steps are two main tasks associated with each step.  For step 1, the main tasks are literature review and workshops and interviews; for step 2, inter-annotator agreement and linguistic analysis; for step 3, feature engineering and error analysis; and for step 4, precision, recall, and F1 score and workshop.  The three dotted line arrows connect the fourth step to the third, second, and first steps to indicate how future work can follow an iterative process of defining a model's task, creating training data, and designing and training models based on the results of both a quantitative and a qualitative evaluation of the models.}
\end{figure*}
    
    Even when the focus of an area of work is to remove bias, the situated nature of any person or ML model carrying out the work means there will always be some bias, as participants' found with the application of the \textbf{\textit{Occupation}} code (O2), as Dias Oliva \textit{et al.} have observed in the silencing of the LGBTQ community as a result of efforts to suppress hate speech online with ML~\cite{DiasOliva_2021}, and as Parmar \textit{et al.} observed with bias rooted in data annotation instructions for crowdworkers~\cite{Parmar_2023}.  Moreover, in the GLAM domain, removing data bias causes harm by erasing evidence of historical injustice.  Records of bias are necessary to learn about equity, fairness, and justice; to hold people and society accountable; and to guide social change~\cite{Blouin_Rosenberg_2011,Pilgrim_2005,SAA}.  ML systems for Information Retrieval (IR) and summarization have similar purposes as GLAM workflows (\textit{i.e.}, describing and providing access to information), thus our findings are especially relevant to how these ML systems are designed.  We encourage IR and summarization research that aims to communicate bias and present more than one perspective to end users, such as news aggregator AllSides' approach to political bias in news reporting.~\footnote{\url{allsides.com}}

    \subsection{RQ2: ML for Informing Understandings and Management of Bias}
    
    From the 15 observations reported in \S\ref{observations}, we identified three ways ML can contribute to information and heritage professionals' understanding and management of bias in GLAM collections and their catalog descriptions:
    \begin{enumerate}
        \item Supporting critical cataloging with estimates of the quantities of potentially biased descriptions per collection, and with the identification of specific descriptions most likely to contain gender biased language, informing the prioritization of collections and collection descriptions to review (O8, O9, O10);
        \item Informing improvements to description practices to avoid perpetuating social biases while staying true to the contents of cultural heritage records (O10, O11); and
        \item Providing evidence of resource needs for cataloging work, including critical cataloging and writing new descriptions (O7, O12).
    \end{enumerate}
    Manual description processes cannot keep pace with the rapid rate at which GLAM acquire cultural heritage (O6, O7), so ML models' ability to process text at large scale can add efficiency to advance information and heritage professionals' primary goal of making cultural heritage discoverable.  Additionally, if a GLAM institution adopts ML models trained on data that were coded by people outside the institution, the models can draw information and heritage professionals' attention to biases that they may not have otherwise considered, expanding their understanding of data bias.  


    \subsection{Limitations and Future Work}\label{limitations}

    Our workshop surfaced concerns of participants relevant to future work combining HCI and ML methods to examine data bias.  We began the workshop by stating that the aim of identifying gender biases in catalog descriptions was to highlight how structural injustice manifests in language and that the aim was not to place the blame or responsibility for gender biased language on participants.  We also reiterated this throughout the workshop.  Nonetheless, participants expressed discomfort with evaluating catalog descriptions for bias, describing the workshop discussion as ``intimidating'' and ``self-revealing.''  This sentiment may have reduced participants' willingness to voice their reflections or questions.  Following D'Ignazio and Klein's data feminism principle to ``elevate emotion and embodiment''~\cite{D_K_2020}, we encourage future work to investigate how to apply human-centered research methods to data bias evaluations that engage domain experts in an empowering manner and minimizes their feelings of discomfort.  Aligning with  Young's concepts of structural injustice and social connection model of responsibility~\cite{Young_2011}, we emphasize that while all those working with data have a responsibility to try to counteract structural injustices reflected and perpetuated in data, they are not solely responsible for, nor are they to blame for, biases in those data or any models trained on them.
    
    Future work will include further error analysis and experiments to improve the text classifiers, and evaluations of their performance on additional data.  Given the higher precision than recall scores for the Omission and Stereotype Classifiers, error analysis will be conducted to investigate whether the difficult-to-classify instances represent one or several manifestations of gender biased language.  Further feature engineering experiments (\textit{e.g.}, including part of speech tags as features) will be conducted with the aim of improving the classifiers' performance, and to further investigate the extent to which lexically and grammatically gendered language correlates with gender biased language.  Running and evaluating the classifiers on datasets in similar and different cultural contexts will provide insight on whether linguistic patterns in bias manifestation hold across different text corpora.  To investigate bias transfer and develop approaches to distinguishing biases from models' pre-training and fine-tuning datasets, future work could include experiments with pre-trained models, comparing the gender biases they classify with those of this paper's models.  When creating ML and AI systems, we encourage researchers and practitioners to consider using traditional ML and our mixed-methods evaluation approach to iterate quickly and at low cost, working to more thoroughly minimize the risk of harm from unknown biases before investing in the resources needed to train deep learning models.  As Figure \ref{fig:future-process} illustrates, the quantitative and qualitative evaluation of models can inform repeated refinements to the conceptualization of a task or problem, the contents and quality of training data, and model design.

\section{Conclusion}\label{conclusion}
While creating unbiased and fair ML systems is a worthy goal, in practice, the contextual, dynamic, and inevitable nature of bias means supposedly fair ML systems may uphold existing, inequitable power relationships in society, risking the perpetuation and amplification of social biases.  ML approaches to bias and fairness must expand to consider when the best aim is to manage, rather than eliminate, bias.  
Through the creation of ML models designed to identify gender biased language, and a human-centered evaluation of the models and their underlying data, we have (a) demonstrated the limitations of predominant ML approaches to bias and fairness, and (b) provided an approach to investigating the feasibility of eliminating bias from an ML system.  We encourage ML researchers and practitioners to use our mixed-methods approach from the start of an ML system's design, so the training dataset and model can be iteratively refined quickly and at low cost until the social biases and their consequences for the ML system's use case are thoroughly understood.

\begin{acks}
Thank you to the archivists, librarians, and curators on the University of Edinburgh's Heritage Collections teams, for their contributions to this research, and to the reviewers, whose thoughtful feedback helped us strengthen this paper.  We also extend our gratitude to the organizations who provided grants to support our research: the University of Edinburgh’s Edinburgh Futures Institute and Informatics Graduate School, and the UKRI EPSRC.
\end{acks}

\bibliographystyle{ACM-Reference-Format}
\bibliography{references}

\appendix

\section{Machine Learning}
Here we detail our machine learning approach for reproducibility.

    \subsection{Data Preprocessing}\label{a:ml-preprocess}
    To preprocess the data we selected for training ML models (\textit{i.e.}, to make the descriptions machine readable), we removed the metadata field names (\textit{e.g.}, ``Title,'' ``Scope and Contents'') from the descriptions and converted the text to lowercase; we kept punctuation and stop words in the dataset.  We tokenized the input data for the token and sequence classification models with NLTK's Punkt tokenizer~\cite{Bird_2002}.  For the sequence classifier, the \textbf{\textit{Feminine}}, \textbf{\textit{Masculine}}, \textbf{\textit{Non-binary}}, \textbf{\textit{Unknown}}, and \textbf{\textit{Occupation}} codes were split into beginning and ending tags for the coded text spans' tokens (\textit{e.g.}, the text span ``Queen Elizabeth'' coded as ``Feminine'' would have a ``B-Feminine'' tag on ``Queen'' and ``I-Feminine'' tag on ''Elizabeth''), as is standard in data preprocessing for named entity recognition.  For the sequence classifier, we grouped the tokens by sentence, added START and END booleans to the beginning and end of each sentence, and included a bias value of 1.0 as a feature, as recommended in scikit-learn crf-suite's documentation~\cite{sklearn-crfsuite}.  For the token and sequence classifiers, we represented the tokens within the descriptions as custom fastText word embeddings~\cite{Bojanowski_2017} of 100 dimensions.\footnote{\url{radimrehurek.com/gensim/models/fasttext.html}}  For the document classifier, we represented complete descriptions as Term Frequency-Inverse Document Frequency (TF-IDF) matrices.\footnote{\url{scikit-learn.org/stable/modules/generated/sklearn.feature_extraction.text.TfidfVectorizer.html}}

    \subsection{Model Design, Training, and Testing}\label{a:ml-design}
    We created three types of models: token, sequence, and document classification models, all trained on the training dataset described in \S\ref{training-dataset}.  The token classification models assigned individual tokens (words and subwords) codes for the gender association of individual tokens, specifically \textit{\textbf{Gendered Pronoun}}, \textit{\textbf{Gendered Role}}, and \textit{\textbf{Generalization}}.  The sequence classification models assigned a running sequence of tokens (\textit{i.e.}, a phrase or other subset of a description) codes for the grammatical or lexical gender associated with a person's name or job title, specifically \textit{\textbf{Feminine}}, \textit{\textbf{Masculine}}, \textbf{\textit{Non-binary}}, \textit{\textbf{Unknown}}, or \textit{\textbf{Occupation}}.  We used sequence instead of token classification for these labels due to the longer length of names and job titles' text spans, which ranged from one to ten tokens.  The document classification models assigned entire descriptions codes for gender biased language, specifically \textit{\textbf{Omission}} and \textit{\textbf{Stereotype}}.  The token and document models were trained in a multilabel task, meaning a model could assign zero, one, or many labels to a token or document.  The sequence models were trained in a multiclass task, meaning a model could assign at most one label to a token, because a token cannot be both a job title and a person, nor should a single instance of a person's name be referred to with terms having multiple grammatical or lexical genders (this does allow for different instances of a person's name to have different grammatical and lexical gender associations, however).
    
    Our code was written in Python.\footnote{\url{www.python.org}}  We used scikit-learn and scikit-learn-based libraries for all the classification models reported in this paper.  The multilabel token classification model is a Classifier Chain\footnote{\url{scikit.ml/api/skmultilearn.problem_transform.cc.html}} of Random Forests,\footnote{\url{scikit-learn.org/stable/modules/generated/sklearn.ensemble.RandomForestClassifier.html}} with the parameter ``random\_state to ``22'' and the remaining parameters set to scikit-learn's defaults.  We refer to this model as the Linguistic Classifier (LC), because it classifies with codes from the taxonomy's Linguistic category: \textbf{\textit{Gendered Pronoun}}, \textbf{\textit{Gendered Role}}, and \textbf{\textit{Generalization}}.  The multiclass sequence classification models are Conditional Random Fields (CRF) models using the AROW algorithm with the parameters ``variance'' set to ``1,'' ``max\_iterations'' set to ``100,'' and ``all\_possible\_transitions'' set to ``True.''  We refer to these model as the Person Name and Occupation Classifiers (PNOCs) because it classifies with codes from the taxonomy's Person Name category (\textbf{\textit{Feminine}}, \textbf{\textit{Masculine}}, and \textbf{\textit{Unknown}}) as well as the \textbf{\textit{Occupation}} code from the taxonomy's Contextual category.  The multilabel document classification models use Support Vector Machines with Stochastic Gradient Descent\footnote{\url{scikit-learn.org/stable/modules/generated/sklearn.linear_model.SGDClassifier.html}} in a one-vs.-rest setup with scikit-learn's default parameters.  We refer to these models as the Omission and Stereotype Classifiers (OSCs) because they classify descriptions \textbf{\textit{Omission}} and \textbf{\textit{Stereotype}} codes from the with the taxonomy's Contextual category.  To arrive at these model setups, we experimented with several algorithms and parameter settings for token, sequence, and document classification, choosing the algorithms and parameters that resulted in the highest F\textsubscript{1} scores (for all experiment detail, see \cite{Havens_2024}).

    To train and test each classification model, we took a modified approach to five-fold cross-validation to generate predictions (model-made codes) for the entire dataset of descriptions.  First, we split the dataset into five subsets, or folds, where each fold accounted for a unique 20\% selection of the dataset.  Then, we iteratively selected four folds (80\% of the dataset) for training and used the remaining fold (20\% of the dataset) for testing.  This approach trained five instances of every model and enabled the generation of predictions for the entire dataset, 20\% at a time.  We evaluated the models using precision, recall, and F\textsubscript{1} scores, optimizing for F\textsubscript{1} score.  These are the same scores calculated to measure the human coders' Inter-Annotator Agreement (IAA) for our training dataset's source data~\cite{Havens_2022}, enabling us to compare the automated and manual coding processes.
    Our code is available on GitHub.\footnote{\url{github.com/thegoose20/gender-bias}}
    We report on the performance of the models using precision, recall and F\textsubscript{1} scores, optimizing for F\textsubscript{1} score.

    \subsection{Model Performance}\label{a:ml-results}
    
    We evaluated each individual text classification model across the three model cascades and report the performance scores per label in tables \ref{t:osc-label-scores}, \ref{t:ling-label-scores}, and \ref{t:pnoc-label-scores}.  Table \ref{t:osc-scores} reports the overall performance scores for the OSCs, the last model in each cascade.  Given space constraints and the focus of our work on leveraging the models to investigate information professionals' conceptualization of gender bias, we give only a brief discussion of the implications of the models' performance scores here (see \cite{Havens_2024} for further detail).
    
    The LC yielded the highest performance scores across all models (Table \ref{t:ling-label-scores}).  This result is not surprising given the high agreement between human coders in the training dataset for two of the codes that the LC classifies tokens with: \textbf{\textit{Gendered Pronoun}} (F\textsubscript{1}=0.957) and \textbf{\textit{Gendered Role}} (F\textsubscript{1}=0.780).  Performance scores for \textbf{\textit{Generalization}}, however, are lowest across all models' per code scores.  Though the model's precision score is reasonably high, at 0.646 (64.6\% of all the model's classified instances were meant to be classified), its recall score is very low, at 0.232 (23.2\% of all instances that were meant to be classified were, in fact, classified by the model), leading to a low (below 50\%) F\textsubscript{1} score of 0.341.  This pattern follows the human coders' agreement with the training dataset, which also had high precision and low recall scores.  These results indicate that there are different language patterns that indicate an instance of gender generalization and a portion of these patterns were not learned or not learned well by the model.  This reflects the confusion that human coders reported during their discussion of how to apply the \textbf{\textit{Generalization}} code throughout the human coding process.

    \begin{table*}[t]
        \caption{Comparing human coders and text classification model's performance for \textbf{\textit{Gendered Pronoun}}, \textbf{\textit{Gendered Role}}, and \textbf{\textit{Generalization}}.  The ``Human Coders'' columns reports the average of the IAA scores between the five human coders for each displayed label.  The ``Coders vs. Training'' column displays the average IAA score of each human coder with the training dataset used for model training.  The ``LC'' columns report the average of the Linguistic Classifier's scores on the test folds during five-fold cross-validation.  Scores are calculated loosely, meaning one code agrees with another code if it exactly matches or overlaps that other code, and both codes have the same label.}\label{t:ling-label-scores}
        \centering\begin{tabular}{@{}llllllllll@{}}
        \toprule
        \multicolumn{1}{l|}{} & \multicolumn{3}{c|}{Coders' IAA} & \multicolumn{3}{c|}{Coders vs. Training} & \multicolumn{3}{c}{LC} \\ \midrule
        \multicolumn{1}{l|}{Label} & Precision & Recall & \multicolumn{1}{r|}{F\textsubscript{1}} & Precision & Recall & \multicolumn{1}{r|}{F\textsubscript{1}}  & Precision & Recall & F\textsubscript{1} \\ \midrule
        \multicolumn{1}{l|}{Gendered Pronoun} & 0.961         & 0.954      & \multicolumn{1}{l|}{0.957}     & 0.986              & 0.956           & \multicolumn{1}{l|}{0.971}         & 0.811 & 0.992 & 0.893 \\
        \multicolumn{1}{l|}{Gendered Role}    & 0.784         & 0.804      & \multicolumn{1}{l|}{0.780}     & 0.803              & 0.833           & \multicolumn{1}{l|}{0.811}         & 0.740 & 0.808 & 0.773 \\
        \multicolumn{1}{l|}{Generalization}   & 0.381         & 0.295      & \multicolumn{1}{l|}{0.277}     & 0.981              & 0.138           & \multicolumn{1}{l|}{0.236}         & 0.646 & 0.232 & 0.341 \\ \bottomrule
        \end{tabular}
    \end{table*}
    \begin{table*}
        \caption{Comparing human coders and text classification models' performance for \textbf{\textit{Feminine}}, \textbf{\textit{Masculine}}, \textbf{\textit{Non-binary}}, \textbf{\textit{Unknown}}, and \textbf{\textit{Occupation}}.  The ``Coders' IAA'' columns reports the average of the IAA scores between the five human coders for each displayed label.  The ``Coders vs. Training'' column displays the average IAA score of each human coder with the training dataset used for model training.  The ``PNOC'' columns report the average of the Person Name and Occupation Classifier's scores on the test folds during five-fold cross-validation.  Scores are calculated loosely, meaning one code agrees with another code if it exactly matches or overlaps that other code, and both codes have the same label.  There were no \textbf{\textit{Non-binary}}-coded text spans in the training dataset so no scores are reported for that code}\label{t:pnoc-label-scores}
        \begin{tabular}{@{}l|lll|lll|lll|lll@{}}
        \toprule
              & \multicolumn{3}{c|}{Coders' IAA} & \multicolumn{3}{c|}{Coders vs. Training} & \multicolumn{3}{c|}{Baseline PNOC} & \multicolumn{3}{c}{Cascade 1 PNOC} \\ \midrule
        Label & Precision  & Recall  & F\textsubscript{1} & Precision     & Recall     & F\textsubscript{1}    & Precision  & Recall  & F\textsubscript{1}  & Precision  & Recall  & F\textsubscript{1}  \\ \midrule
        Feminine   & 0.604 & 0.601 & 0.597 & 0.992 & 0.682 & 0.807 & 0.636 & 0.646 & 0.641 & 0.487 & 0.675 & 0.566 \\
        Masculine  & 0.686 & 0.667 & 0.665 & 0.997 & 0.696 & 0.817 & 0.506 & 0.344 & 0.409 & 0.415 & 0.398 & 0.406 \\
        Non-binary  & -     & -     & -     & -     & -     & -     & -     & -     & -     & -     & -     & -     \\
        Unknown    & 0.677 & 0.725 & 0.678 & 0.996 & 0.800 & 0.879 & 0.589 & 0.636 & 0.611 & 0.591 & 0.464 & 0.520 \\
        Occupation & 0.660 & 0.749 & 0.698 & 0.934 & 0.809 & 0.866 & 0.657 & 0.599 & 0.627 & 0.652 & 0.569 & 0.608 \\ \bottomrule
        \end{tabular}
    \end{table*}

    The PNOCs yielded low per code scores relative to the LC and OSCs, with F\textsubscript{1} scores ranging from 0.406 to 0.641 (Table \ref{t:pnoc-label-scores}).  Despite the greater presence of men in the archival catalog and the collections it documents, the \textbf{\textit{Masculine}} code had the lowest score amongst all PNOC codes.  Error analysis on the PNOC and human coders' classified data found that \textbf{\textit{Masculine}} and \textbf{\textit{Unknown}} labels were frequently confused.  This indicates a high difficulty of the human coding task: though the taxonomy instructed human coders to label names without a grammatically or lexically gendered referent (\textit{e.g.}, ``he'' or ``Sir'') as \textbf{\textit{Unknown}}, many person names without such a referent were coded as \textbf{\textit{Masculine}}.  This could be due to previous descriptions a human coder had read about the same person that had a grammatically or lexically masculine referent or due to stereotypical inferences made about a person based on their name or other information provided about them in a description, such as their job title.  The human coders did not find any instances of grammatically or lexically non-binary referents, resulting in no \textbf{\textit{Non-binary}} codes in the training dataset for models to learn from; this is thus a gap in our model that can be addressed in future work by expanding our dataset.  Note that there may still be people in our dataset who identified as non-binary or another gender diverse identity.  Due to the marginalization and oppression of people of non-binary and gender diverse identities, people may have chosen not to document or to destroy records that documented their gender identity~\cite{Shopland_2020}.

    Among the three cascades' OSCs, the OSC for cascades 2 and 3 yielded the highest per label scores.  Cascade 3's OSC had the highest F\textsubscript{1} score for \textbf{\textit{Omission}}, at 0.682, which was slightly lower than the human coders' agreement with the aggregated (training) dataset at (F\textsubscript{1}=0.767) and much higher than the human coders' IAA (F\textsubscript{1}=0.439).  Cascade 2's OSC had the highest \textbf{\textit{Stereotype}} F\textsubscript{1} score at 0.841, which is higher than the human coders' IAA (F\textsubscript{1}=0.440) and human coders' agreement with the training dataset (F\textsubscript{1}=0.823).  Across all labels, Cascade 3's OSC yielded the highest F\textsubscript{1} scores (macro F\textsubscript{1}=0.761, micro F\textsubscript{1}=0.730), though Cascade 2's OSC was a close second (macro F\textsubscript{1}=0.760, micro F\textsubscript{1}=0.727).  Given the better performance scores of the LC relative to the PNOC, we posit that Cascade 2's OSC is more reliable, because Cascade 2 included the LC's codes as features while Cascade 3 included the PNOC's codes as features.  Surprisingly, the inclusion of both the LC and PNOC's codes as features for Cascade 1 yielded the lowest OSC performance, with this OSC yielding macro and micro F\textsubscript{1} scores within 0.001 of the baseline OSC.  We also note that the inclusion of taxonomy codes reduced the precision scores (the baseline OSC's precision scores were highest) but improved the recall scores.  There was a greater improvement to recall than reduction in precision, however, leading to higher overall OSC performance as measured with macro and micro F\textsubscript{1} scores.

\end{document}